\documentclass{article}
\usepackage{arxiv}

\usepackage{amsmath,amsfonts,bm}

\def\eqref#1{equation~\ref{#1}}

\def\1{\bm{1}}

\DeclareMathAlphabet{\mathsfit}{\encodingdefault}{\sfdefault}{m}{sl}
\SetMathAlphabet{\mathsfit}{bold}{\encodingdefault}{\sfdefault}{bx}{n}

\definecolor{scholarblue}{rgb}{0.21,0.49,0.74}
\definecolor{darkblue}{rgb}{0, 0, 0.5}
\usepackage[pagebackref,breaklinks,colorlinks, citecolor=scholarblue]{hyperref}

\usepackage[utf8]{inputenc}
\usepackage[T1]{fontenc}
\usepackage{hyperref}
\usepackage{url}
\usepackage{booktabs}
\usepackage{amsfonts}
\usepackage{nicefrac}
\usepackage{microtype}
\usepackage{xcolor}
\usepackage{graphicx}

\usepackage[most]{tcolorbox}
\usepackage{tabularx}
\usepackage{multirow}
\usepackage{adjustbox}
\usepackage{colortbl}
\usepackage{comment}
\usepackage{subcaption}
\usepackage{float}
\usepackage[most]{tcolorbox}
\usepackage{amsmath}
\usepackage{pifont}
\usepackage{xspace}
\usepackage{wrapfig}
\usepackage{enumitem}
\usepackage{comment}
\usepackage[textsize=tiny]{todonotes}

\usepackage{hyperref}
\usepackage{url}
\usepackage{graphicx}
\usepackage{booktabs}
\usepackage{amsmath,amssymb,amsthm}
\usepackage{dsfont}
\usepackage{bbm}
\usepackage{mathtools}
\usepackage{xcolor}
\usepackage{multirow}
\usepackage{xspace}
\usepackage{enumitem}
\usepackage{microtype}
\usepackage[most]{tcolorbox}
\usepackage[capitalize,noabbrev]{cleveref}
\usepackage{pifont}
\usepackage{subcaption}
\usepackage{overpic}
\usepackage{array}
\usepackage{tikz}
\usepackage{graphicx}
\usetikzlibrary{positioning,calc,arrows.meta,decorations.pathreplacing}

\definecolor{ctxcol}{HTML}{F5A83D}
\definecolor{intentcol}{HTML}{729CFE}
\definecolor{questioncol}{HTML}{88B06D}
\definecolor{anscol}{HTML}{F65550}

\definecolor{ctxanscol}{HTML}{F5433D}
\definecolor{prianscol}{HTML}{5D8DFD}

\definecolor{viscol}{HTML}{897bc1}
\definecolor{txtcol}{HTML}{F5A83D}
\definecolor{priorcol}{HTML}{F65550}

\definecolor{accent}{HTML}{0064E0}

\newcommand{\ctxtext}[1]{{\color{ctxcol}\textit{#1}}}

\newcommand{\questiontext}[1]{{\color{questioncol}\textit{#1}}}
\newcommand{\anstext}[1]{{\color{anscol}\textit{#1}}}

\newcommand{\benchmark}{\textsc{WhatIfVis}\xspace}
\newcommand{\ccs}{controllable context sensitivity\xspace}
\newcommand{\CCS}{Controllable Context Sensitivity\xspace}
\newcommand{\cvcs}{controllable visual context sensitivity\xspace}

\newcommand{\vctx}{\textcolor{viscol}{\textbf{visual context}}\xspace}
\newcommand{\tctx}{\textcolor{txtcol}{\textbf{textual context}}\xspace}
\newcommand{\firstH}{\textbf{H1}\xspace}
\newcommand{\secH}{\textbf{H2}\xspace}

\newcommand{\lm}{p}
\newcommand{\alphabet}{\Sigma}
\newcommand{\question}{{\color{questioncol}{\boldsymbol{q}}}}

\newcommand{\queries}{{\color{questioncol}{\mathcal{Q}}}}
\newcommand{\context}{{\color{ctxcol}{\boldsymbol{c}}}}

\newcommand{\contexts}{{\color{ctxcol}{\mathcal{C}}}}
\newcommand{\answer}{{\color{anscol}{a}}}

\newcommand{\answerstr}{{\color{anscol}\boldsymbol{a}}}
\newcommand{\answerprior}{\answer(\question, \varepsilon)}
\newcommand{\answerctx}{\answer(\question, \context)}
\newcommand{\ctxweight}{{\color{intentcol}{w}}}
\newcommand{\ctxintent}{{\color{intentcol}{\mathrm{ctx}}}}
\newcommand{\priorintent}{{\color{intentcol}{\mathrm{pri}}}}
\newcommand{\ctxweightdomain}{\left\{\ctxintent, \priorintent\right\}}
\newcommand{\formatprompt}{F}
\newcommand{\pairacc}{\ensuremath{\mathrm{PairAcc}}}
\newcommand{\QCtrn}{\mathcal{S}_{\text{trn}}}
\newcommand{\QCtst}{\mathcal{S}_{\text{tst}}}

\newcommand{\subspace}{\mathcal{F}}

\newcommand{\uvec}{\boldsymbol{u}}

\newcommand{\cfunc}{c}
\newcommand{\knob}{\ensuremath{\cfunc(\ctxweight)}}

\newcommand{\image}{{\color{viscol}{\mathbf{I}}}}
\newcommand{\statement}{{\color{ctxcol}{s}}}
\newcommand{\evidence}{{\color{ctxcol}{e}}}
\DeclareMathOperator*{\greedy}{greedy}

\theoremstyle{definition}

\newtcolorbox{takeaway}{
  colback=blue!5,
  colframe=blue!70!black,
  arc=5pt,
  boxsep=5pt,
  left=2pt, right=2pt, top=2pt, bottom=2pt,
  boxrule=0.8pt,
  drop shadow=gray!50!white,
  enhanced jigsaw,
  before skip=12pt plus 2pt minus 2pt,
  after skip=12pt plus 2pt minus 2pt,
}
\newcommand{\takeawayhead}{\textbf{\textit{Takeaway:}}\xspace}
\newcommand{\rparagraph}[1]{\noindent\textbf{#1}~~}

\definecolor{mintgreen}{RGB}{152, 255, 152}
\makeatletter
\newcommand*\iftodonotes{\if@todonotes@disabled\expandafter\@secondoftwo\else\expandafter\@firstoftwo\fi}
\makeatother

\title{Seeing or Knowing? Visual Context Sensitivity in Multimodal Large Language Models}

\author{
\textbf{Jiaang Li}$^{1}$ \quad
\textbf{Chengzu Li}$^{1,2}$ \quad
\textbf{Zhaochong An}$^{1}$ \quad
\textbf{Yifei Yuan}$^3$ \\
\textbf{Xi Liu}$^4$ \quad
\textbf{Serge Belongie}$^1$ \quad
\textbf{Vésteinn Snæbjarnarson}$^1$ \\[0.9em]
$^1$University of Copenhagen \quad $^2$University of Cambridge \\
$^3$ETH Zürich \quad
$^4$Independent
}

\begin{document}

\maketitle

\begin{abstract}
Multimodal Large Language Models (MLLMs) achieve strong performance by integrating visual inputs with the rich priors of pretrained language models. However, they often fail on vision-centric tasks, especially when visual evidence conflicts with pretrained knowledge. We explore these failures separately using two diagnostic paradigms: (1) probing whether visual information is available, via image reconstruction, and (2) measuring multimodal context sensitivity, the extent to which the model follows visual context versus the language prior. To support the second, we introduce the \benchmark, a benchmark spanning five coarse-grained dimensions (spatial-temporal, color, count, size, and weight) whose questions admit answers from either the image or the prior.
Our analysis yields three findings:
(i) Coarse-grained visual evidence is preserved, as these attributes can be reconstructed from the final-layer image tokens of frozen MLLMs. Failures on questions about these attributes therefore point to post-perceptual utilization, rather than to degraded visual encoding during perception.
(ii) Even when explicitly instructed to use or ignore visual evidence, vanilla models (without supervised fine-tuning on the \benchmark) show unstable visual context sensitivity. Supervised fine-tuning (SFT) improves this controllability and generalizes across domains, and activation patching further localizes the vision‑versus‑prior trade-off at architecture‑specific depths across all six models.
(iii) The vision-versus-prior trade-off is controllable along a learned vector. Applying this steering vector, even without any intent instruction, improves controllability over the vanilla model.
Together, these results relocate the bottleneck, indicating that for the coarse attributes we study, MLLMs encode the visual evidence but cannot reliably control their reliance on it.\footnote{Code: \href{https://github.com/jiaangli/visual-context-sensitivity}{\textcolor{scholarblue}{visual-context-sensitivity}}}
\end{abstract}

\section{Introduction}
\label{sec:intro}
Multimodal large language models (MLLMs) derive a substantial portion of their capabilities from the underlying large language models (LLMs)~\citep{laurencon2024building}.
Pretraining on trillions of text tokens endows the LLM backbone with broad world knowledge and commonsense reasoning, providing the foundation on which MLLMs power diverse real-world applications by integrating what they see (image context) with what they know (prior parametric knowledge). Yet the same prior that grants this competence can also betray it: grown confident enough, it overrides what the vision encoder "sees".
Helmholtz cast perception itself as inference under prior beliefs, in which the senses propose and the prior disposes, whose signature failure is to register what one expects rather than what is present~\citep{helmholtz1867handbuch}.

We suspect an MLLM inherits the same fault line by construction. The LLM backbone is pretrained on far more text than the paired image-text data used for vision-language alignment, leading the MLLM to rely heavily on its prior world knowledge when answering questions about an image. As a result, when presented with an image depicting something outside its prior expectations, the model may favor its learned prior knowledge over the visual evidence.

\begin{figure}
\centering
\includegraphics[width=1\linewidth]{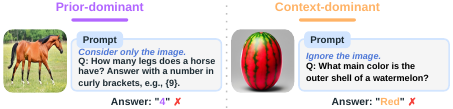}
\caption{\textbf{One MLLM, two opposite failures.} A single model lacks a stable setpoint on the visual context-sensitivity axis between its parametric \textcolor{viscol}{\textbf{prior}} and the \textcolor{txtcol}{\textbf{image}}, giving rise to two opposite failure modes: (Left) Prior-dominant failures, where strong parametric priors overwhelm anomalous visual input; and (Right) Context-dominant failures, where the model over-indexes on visual features even when instructed to ignore them.}
\label{fig:teaser}
\end{figure}
Consistent with this hypothesis, recent work shows that, across six frontier models, MLLMs frequently report what they \emph{know} rather than what they \emph{see} \citep{vo2026vision} (Figure~\ref{fig:teaser}, left).
The field's reflex is to blame perception, arguing vision encoders are known to discard detail~\citep{tong2024eyes}.
MLLMs inherit the representational limits of those encoders~\citep{tong2024cambrian,tang2026unilip}, and combining or enhancing vision encoders is the standard remedy~\citep{tong2024cambrian,liu2025tuna,liu2026tuna}.
On this account the visual evidence is degraded before an answer is formed, and the model falls back on its prior.
However, as shown in Figure~\ref{fig:teaser} (right), the failure can also run in reverse.
When presented with a watermelon whose rind is red and explicitly asked for its \emph{usual} color, the model still focuses on the image and doesn't retrieve the canonical \emph{green}; it remains unable to suppress the visual evidence even under direct instruction. A merely degraded signal cannot explain a model that follows the image against an explicit instruction. The deficit is therefore not that MLLMs ignore the image or over-trust it, but that they use the it \emph{unreliably}. Fundamentally, this is a issue of \emph{visual context sensitivity}: models have no consistent, controllable policy (e.g., instructions) for when to use the visual evidence and when to override it, especially when visual evidence and its parametric prior conflicts.

We frame the deficit as two hypotheses:
\begin{itemize}
    \item \textbf{\firstH: Perception failure.} The vision channel transmits too little, and the visual evidence is degraded by the time an answer is formed and the model falls back on its prior.
    \item \textbf{\secH: Utilization failure.} The visual evidence is present in the LLM backbone, yet the model has no consistent, controllable policy (e.g., instructions) for when to heed it and when to override it.
\end{itemize}

While standard multimodal benchmarks~\citep{yue2024mmmu,liu2024mmbench,li2026ravenea} effectively measure end-to-end performance, they conflate perception with utilization. Because perception is a strict prerequisite for utilization, as a model cannot leverage visual evidence unless it has first perceived it, a downstream failure can stem from either a visual blind spot or a utilization error. By bundling these two distinct stages into a single score, existing suites fail to isolate \firstH.

In this work, we explicitly disentangle \firstH to demonstrate that \secH is a genuine, separable failure and turn it into a means of control. To isolate \firstH, we focus on coarse object-level attributes (e.g., identity, color, size, count, and weight) that vision encoders capture reliably, rather than fine-grained information such as textures or small objects, which encoders tend to discard during encoding~\citep{wu2024vstar}. We then evaluate visual evidence in a unified manner by measuring how faithfully the input image can be reconstructed solely from the \emph{final-layer image tokens} of frozen MLLMs. Our results show that, for \emph{the coarse attributes} we probe, the visual evidence is \textbf{preserved} in the final-layer image tokens, indicating that \emph{the observed failures stem from utilization rather than perception}.
We next investigate whether \secH can be alleviated by better controlling visual context sensitivity through a one‑dimensional subspace, a knob that modulates reliance on visual context. To this end, we adapt the \ccs (CCS) framework~\citep{minder2025controllable} to visual contexts and term this extension \cvcs (CVCS). To evaluate \secH, we introduce the \benchmark benchmark, whose questions are deliberately designed so that the answer changes depending on whether the model relies on the image or on its prior knowledge (Figure~\ref{fig:overview_dataset} and \ref{fig:benchmark_app}). We find that vanilla models, when evaluated with a strict conjunctive metric demanding simultaneous adherence to the image‑reliance intent and the prior‑reliance intent, yield too few successful examples to reliably localize the vision-versus-prior knobs. We therefore localize the knobs in the supervised fine-tuned (SFT) models and show that they can be used directly in the corresponding vanilla models, substantially improving control over visual context sensitivity.

Our contributions and key findings include:
\begin{itemize}
\item \textbf{A generative probe for visual information.} To isolate \firstH, we reconstruct counterfactual images directly from the final-layer image tokens of frozen MLLMs. We find coarse counterfactual content remains encoded in the final layer across three models from different families. This finding indicates that failures on these attributes stem from how visual evidence is \emph{used}, rather than from perception (Section~\ref{sec:reconstruction}).
\item \textbf{A benchmark for bidirectional context sensitivity.} We introduce the \benchmark benchmark, which curates counterfactual images from existing benchmarks under matched textual controls and introduces two strict conjunctive \pairacc\ metrics. Evaluating vanilla models on this benchmark reveals \secH, where models over-trust visual evidence on some attributes yet disregard it on others, a brittleness hidden by conventional marginal accuracy metrics (Section~\ref{sec:knob}).
\item \textbf{Per-model scalar knobs for controllable visual context sensitivity.} Using activation patching, we identify the layers responsible for resolving the trade-off between visual evidence and prior knowledge, and uncover a compact one-dimensional subspace in the model's final responsible layer. This subspace serves as an effective knob that modulates the model's reliance on visual context versus prior knowledge. Manipulating this knob improves controllability over the vanilla model, even without explicit intent instructions (Section~\ref{sec:steering}).

\item \textbf{Generality across architectures and modalities.} We show that the vision-versus-prior trade-off resolves in a narrow band of a few layers in all six models we examined, yet at architecture-specific depths. These models span three families with distinct attention designs, namely a standard transformer (Qwen2.5-VL~\citep{bai2025qwen25vltechnicalreport}), a linear-attention hybrid (Qwen3.5~\citep{qwen3.5}), and a per-layer-embedding design (Gemma-4~\citep{gemmateam2026gemma4}) (Section~\ref{sec:knob}). We also uncover an asymmetry where text-based control is far more reliable than vision-based control, a gap that widens with model scale (Section~\ref{sec:modal_comparison}).
\end{itemize}

\section{Background}
\label{sec:background}
\rparagraph{Multimodal language models.}
The dominant recipe for building an MLLM couples a pretrained LLM $p_\theta$ over a token vocabulary $\Sigma$ with a \textbf{vision encoder} $\mathcal{E}$ through a lightweight \textbf{projector} $\mathcal{P}$ \citep{liu2023visual,liu2024improved}.
Given an image $\mathbf{V}$, the encoder produces a sequence of patch features $\mathcal{E}(\mathbf{V}) = (\mathbf{v}_1, \dots, \mathbf{v}_K) \in \mathbb{R}^{K \times d_v}$; the projector maps each into the language model's embedding space, yielding \textbf{visual tokens} $ \mathcal{P}(\mathcal{E}(\mathbf{V}))=(\mathcal{P}(\mathbf{v}_1), \dots, \mathcal{P}(\mathbf{v}_K)) \in \mathbb{R}^{K \times d}$.
The MLLM then conditions on the concatenation of visual and text tokens autoregressively:

\begin{equation}
p_\theta\!\left(y_t \mid \mathcal{P}(\mathcal{E}(\mathbf{V})),\, x_{<t}\right)
= \mathrm{softmax}\!\left(\mathbf{W}\, \mathbf{h}_t^{(L)}\right)_{\!y_t},
\label{eq:mllm}
\end{equation}

where $\mathbf{h}_t^{(L)} \in \mathbb{R}^D$ is the residual-stream vector at position $t$ after the final transformer layer $L$ and $\mathbf{W}$ is the unembedding matrix.
Vision encoders are typically language-supervised vision transformers, CLIP \citep{radford2021learning} or SigLIP \citep{zhai2023sigmoid,tschannen2025siglip}, and projectors range from MLP layers to resamplers \citep{liu2023visual,li2023blip,alayrac2022flamingo}.
Our study uses three families that instantiate this recipe: Qwen2.5-VL \citep{bai2025qwen25vltechnicalreport}, Qwen3.5 \citep{qwen3.5}, and Gemma-4 \citep{gemmateam2026gemma4}.

\rparagraph{Subspace intervention.}
In text setting, \citet{minder2025controllable} shows the instruction of using context is encoded in a vector in the residual stream. Consequently, once this vector is identified, the model's context sensitivity can be controlled by editing it, while leaving the remaining information largely unchanged.
Mathematically,
let $\mathbf{h}_t^\ell \in \mathbb{R}^D$ denote the residual-stream vector at the last token position after a critical layer $\ell$ identified by the patching search.
We learn a unit vector $\mathbf{u} \in \mathbb{R}^D$ ($\|\mathbf{u}\|=1$) and form the \textbf{rank-1 orthogonal projection} $P = \mathbf{u}\mathbf{u}^\top$, which decomposes the representation into an intent-encoding component and an orthogonal complement:

\begin{align}
\mathbf{h}_t^\ell &= (I - P)\,\mathbf{h}_t^\ell \;+\; P\,\mathbf{h}_t^\ell, \label{eq:decomp} \\
\tilde{\mathbf{h}}_t^\ell &= (I - P)\,\mathbf{h}_t^\ell \;+\; P\,\mathbf{h}_s^\ell, \label{eq:patch}
\end{align}

where $I$ is the identity matrix.
Equation~\ref{eq:decomp} is the trivial decomposition and equation~\ref{eq:patch} replaces only the component along $\mathbf{u}$ with that of a source activation $\mathbf{h}_s^\ell$, leaving all information orthogonal to intent intact.
We learn $\mathbf{u}$ by minimizing the cross-entropy of predicting the source answer when $\tilde{\mathbf{h}}_t^\ell$ is fed to the remaining layers, with all other model parameters frozen.

\rparagraph{Static steering.}
Once $\mathbf{u}$ is learned, intent can be set \emph{without} any instruction in the prompt by clamping the subspace value to a scalar constant:

\begin{equation}
\tilde{\mathbf{h}}_t^\ell = (I - P)\,\mathbf{h}_t^\ell \;+\; \mathbf{u}\, c(w),
\label{eq:steer}
\end{equation}

where $c(w) \in \mathbb{R}$ is a model-specific scalar for each intent $w$(one value to follow the context, another to follow the prior). Whether such a knob also exists for \emph{visual} context is the question we take up in Sections~\ref{sec:knob}--\ref{sec:steering}.

\section{Image Reconstruction from MLLMs}
\label{sec:reconstruction}
Prior work has largely settled the \emph{fine-grained} case, demonstrating that vision encoders provably discard high-frequency details such as precise textures and small objects~\citep{tong2024eyes,wu2024vstar}. This leaves the natural question open at the other end of the spectrum: what about \emph{coarse-grained} content (e.g., count, color, and so on) that a competent encoder should have no trouble encoding?
If the content is degraded before the LLM backbone can use it, the failure is \emph{\firstH}; if it survives yet the model still fails, the limit is \emph{\secH}. We probe exactly the regime, and show that coarse counterfactual content is present at the final layer of MLLMs.

Downstream probing (attaching different task heads and fine-tuning)~\citep{pang2026unveiling} and multiple-choice MLLM evaluation suites~\citep{liu2024mmbench,yue2024mmmu,li2026ravenea,li-etal-2024-foodieqa} both measure task \emph{performance}, which conflates the two hypotheses we need to separate: a low score is equally consistent with information never arriving (\firstH) and with information arriving but going unused (\secH). Worse, both let the language prior supply the answer, so a model can score well without reading the image at all; the very failure we are diagnosing. Testing \firstH, however, requires a measurement of \emph{what visual content is present in the backbone's representation}, independent of whether the model chooses to act on it. This is precisely what standard evaluations cannot provide. We therefore evaluate the representations \textit{generatively}, attempting to reconstruct images from the LLMs' backbone representations rather than simply scoring their answers. If the counterfactual content can be redrawn from those representations, it was demonstrably present.

\rparagraph{Probing Design via Reconstruction.}
We reconstruct images \textit{solely from final-layer image tokens} of frozen MLLMs.
We adapt the framework of Metaquery~\citep{pan2025transfer} to \textit{non-learnable} queries, and pair a frozen pretrained MLLM, with a trainable 6-layer bidirectional transformer connector and an image decoder SANA~\citep{xie2025sana}. To avoid the leakage of the counterfactual information, we train the connector and decoder with only the real world images. Since the connector and decoder never see counterfactuals in training, the recovery is a faithful readout of what the final-layer representation retains, not a decoder manufacturing the attribute from its prior. In contrast to prior reconstruction-based approaches such as UniLIP~\citep{tang2026unilip} and UniPic2~\citep{wei2025skyworkunipic20building}, which additionally leverage VAE~\citep{kingma2013auto} image tokens to improve generation quality, our objective is not photorealistic image synthesis but rather probing the extent to which visual representations are preserved inside MLLMs. Consequently, we discard learnable question tokens and reconstruct images solely from the final-layer image tokens from the MLLM.
\begin{figure}
    \centering
    \includegraphics[width=1\linewidth]{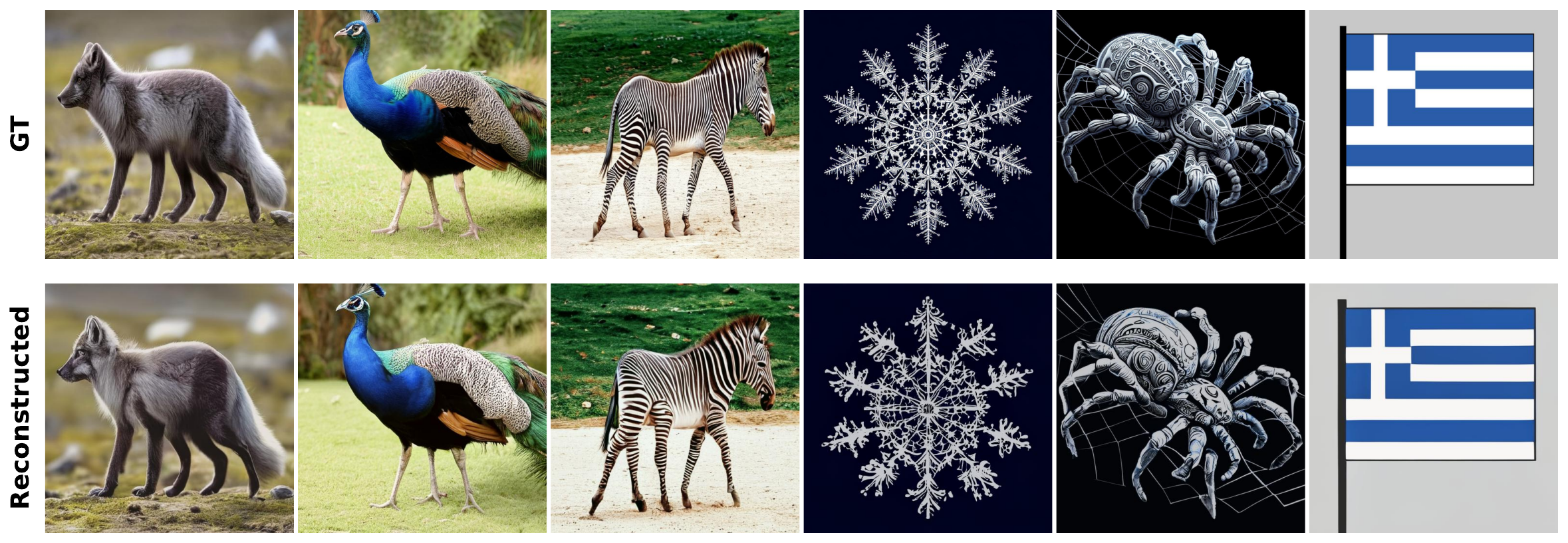}
    \caption{\textbf{The counterfactual attribute survives to the final layer and is visible in the reconstruction.} Species identity, pose, structure, and dominant colors are preserved, including the counterfactual attribute (e.g.\ the anomalous limb count), while high-frequency texture softens.}
    \label{fig:recon}
\end{figure}
We do the cross-reference evaluation on the animal-category subset of VLMs-are-Biased~\citep{vo2026vision}. Three MLLMs (Qwen2.5VL-3B~\citep{bai2025qwen25vltechnicalreport}, Qwen3.5-4B~\citep{qwen3.5} and Gemma-4-E2B-IT~\citep{gemmateam2026gemma4}) all fail in this set with near zero performance. The connector and the decoder, trained on ImageNet-1K~\citep{5206848}, render animals faithfully enough to expose whether the counterfactual content was retained. See the extra results for more subsets in Table~\ref{tab:recon_attributes}.
\begin{table}[t]
\centering
\begin{minipage}[t]{0.4\textwidth}
\centering
\caption{\textbf{Human evaluation of reconstructed images.}
Across three LLM backbones, the counterfactual attribute (count) is preserved in the vast majority of generated samples. CF denotes counterfactual.}
\label{tab:recon_human}
\resizebox{\linewidth}{!}{
\begin{tabular}{lcc}
\toprule
\textbf{Backbone}  & \textbf{CF(\%)} & \textbf{Real(\%)} \\
\midrule
Gemma-4-E2B-IT & \cellcolor{cyan!5}\textbf{97.8} & 2.2  \\
Qwen2.5VL-3B  & \cellcolor{cyan!5}\textbf{93.4} & 6.6 \\
Qwen3.5-4B  & \cellcolor{cyan!5}\textbf{97.8} & 2.2 \\
\midrule
\end{tabular}
}
\end{minipage}
\hfill
\begin{minipage}[t]{0.55\textwidth}
\centering
\caption{\textbf{The counterfactual evidence (count) is preserved in the final layer.}
Reconstructions from frozen MLLMs are metrically closer to the counterfactual reference than to the real-world image across all models and metrics. Ref. = reference; cf = counterfactual; real = real-world.}
\label{tab:recon_metrics}
\resizebox{\linewidth}{!}{
\begin{tabular}{@{}llcccc@{}}
\toprule
\textbf{Backbone} & \textbf{Ref.} & \textbf{mIoU$\uparrow$} & \textbf{PSNR$\uparrow$} & \textbf{SSIM$\uparrow$} & \textbf{LPIPS$\downarrow$} \\
\midrule
\multirow{2}{*}{Gemma-4-E2B-IT} & real & 0.83 & 15.4 & 0.43 & 0.33\\
 & \cellcolor{cyan!5}cf & \cellcolor{cyan!5}\textbf{0.94} & \cellcolor{cyan!5}\textbf{17.9} & \cellcolor{cyan!5}\textbf{0.52} & \cellcolor{cyan!5}\textbf{0.23} \\
\midrule
\multirow{2}{*}{Qwen2.5VL-3B} & real & 0.78 & 15.2 & 0.44 & 0.35 \\
 & \cellcolor{cyan!5}cf  & \cellcolor{cyan!5}\textbf{0.86} & \cellcolor{cyan!5}\textbf{16.2} & \cellcolor{cyan!5}\textbf{0.48} & \cellcolor{cyan!5}\textbf{0.28} \\
\midrule
\multirow{2}{*}{Qwen3.5-4B} & real & 0.79 & 15.4 & 0.43 & 0.33\\
 & \cellcolor{cyan!5}cf & \cellcolor{cyan!5}\textbf{0.88} & \cellcolor{cyan!5}\textbf{16.6} & \cellcolor{cyan!5}\textbf{0.47} & \cellcolor{cyan!5}\textbf{0.27} \\
\bottomrule
\end{tabular}
}
\end{minipage}
\end{table}

\rparagraph{Results.}
The human evaluation is designed to ask whether the \emph{counterfactual attribute} survives. Two annotators viewed each reconstruction and answered the object, its color, and its leg count, scored against the ground truth (inter-annotator agreement Cohen's $\kappa{=}0.84$). Among three models, reconstructions from counterfactual tokens recover the anomalous count correctly $93.4$, $97.8$, $97.8$ percent, respectively (Table~\ref{tab:recon_human}). The decoder reproduces the anomalous attribute only when the tokens carry it. This finding is further consolidated with the pixel-level cross-reference. We compare each reconstruction against both its \emph{own} counterfactual reference and the paired natural image along four complementary metrics, and on every one it is closest to the counterfactual (Table~\ref{tab:recon_metrics}). The metrics probe different scales of fidelity. mIoU, computed over SAM3 masks~\citep{carion2026sam}, asks whether the \emph{same objects occupy the same regions and shapes}. Its high value (over 0.86) shows that identity, boundaries, and layout survive the entire backbone intact, which is exactly the coarse structure our conflicts turn on. The consistently moderate PSNR, SSIM, and LPIPS scores reveal a clear trade-off, where global structure is preserved yet fine pixel-level details, especially high-frequency textures, are poorly reproduced. Qualitatively, species identity, compositional structure, and the counterfactual attribute are preserved, with high-frequency texture the primary degradation (Figure~\ref{fig:recon}).
\begin{takeaway}
\takeawayhead At the coarse-grained level we probe, the counterfactual evidence survives in the LLM backbones of MLLMs. It is present and decodable from the final-layer image tokens of three architecturally distinct MLLMs, which shows the content is \emph{present}, but the model cannot or does not use it.
\end{takeaway}

\section{Finding the Knob Behind Visual Context Sensitivity}
\label{sec:knob}
\CCS (CCS) was formulated for text-only conflicts \citep{minder2025controllable}, where the conflicting context already resides in the text token stream as a sentence. In our \cvcs (CVCS) setting, by contrast, the conflicting evidence is presented visually and must first be preserved in the MLLMs. We thus first apply the presence test developed in Section~\ref{sec:reconstruction}. Given presence, we evaluate controllability: whether the model can be steered to use or disregard the visual counterfactual on demand.

First, we evaluate vanilla models and adapt them via SFT to establish that visual controllability is learnable and transferable across tasks. Second, we use activation patching on the SFT models to localize the vision-versus-prior trade-off to a narrow layer band. Third, we replace the intent instructions entirely with a subspace steering intervention (Section~\ref{sec:steering}).
\begin{figure}
    \centering
    \includegraphics[width=1\linewidth]{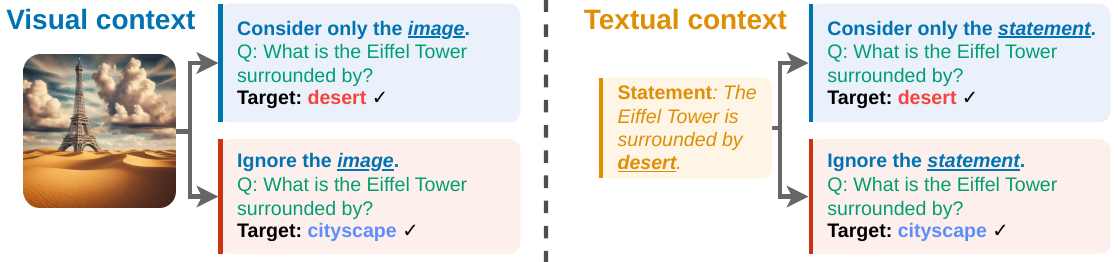}
    \caption{\textbf{Evaluation pipeline for controllable multimodal context sensitivity.} We probe a model's sensitivity to visual and textual contexts separately through targeted instruction following. Given a visual (left) or textual (right) context, the model must answer a question by either strictly grounding its reasoning in the explicit context ("Consider only...") or suppressing the context to recall standard world knowledge ("Ignore...").
    }
    \label{fig:overview_dataset}
\end{figure}
\subsection{Designing the Multimodal Context-Sensitivity Task}
\label{sec:formulation}
Following the text-only setup~\citep{minder2025controllable}, we construct minimal multimodal pairs where the visual evidence and the question remain identical; the only difference is whether the model is instructed to ground its answer in the \textit{given visual evidence} or to rely on its own \textit{prior knowledge}. Holding all other factors constant, these paired examples provide a controlled setting for comparing the model’s internal representations when using visual evidence versus prior knowledge.

Consider an MLLM $\lm$ over an alphabet $\alphabet$. Let $\queries$ be a set of licit questions and $\contexts$ a set of evidence, where each piece of evidence $\evidence \in \contexts$ is either a counterfactual image $\image$ (\vctx) or a one-sentence statement $\statement$ (\tctx) asserting the same content. For a question $\question$ (e.g., \questiontext{What is the Eiffel Tower surrounded by?}) and evidence $\evidence$ (e.g., an image showing \ctxtext{the Eiffel Tower surrounded by desert}), let $\answerprior$ be the context-independent answer (\anstext{cityscape}) and $\answerctx$ the context-dependent one (\anstext{desert}). An \emph{intent} $\ctxweight \in \ctxweightdomain$ selects whether to follow the evidence ($\ctxintent$, target $\answerctx$) or the prior ($\priorintent$, target $\answerprior$), realized by a formatting function $\formatprompt$ that maps $(\question,\evidence,\ctxweight)$ to a prompt. On a training split $\QCtrn$, models see both $\formatprompt(\question, \evidence, \priorintent)\cdot\answerprior$ and $\formatprompt(\question, \evidence, \ctxintent)\cdot\answerctx$ ($\cdot$ = concatenation); a disjoint $\QCtst$ is held out for testing. Two modalities examples are shown in Figure~\ref{fig:overview_dataset}.

\subsection{The \benchmark Counterfactual Benchmark}
\label{sec:benchmark}
\begin{figure}
    \centering
    \includegraphics[width=1\linewidth]{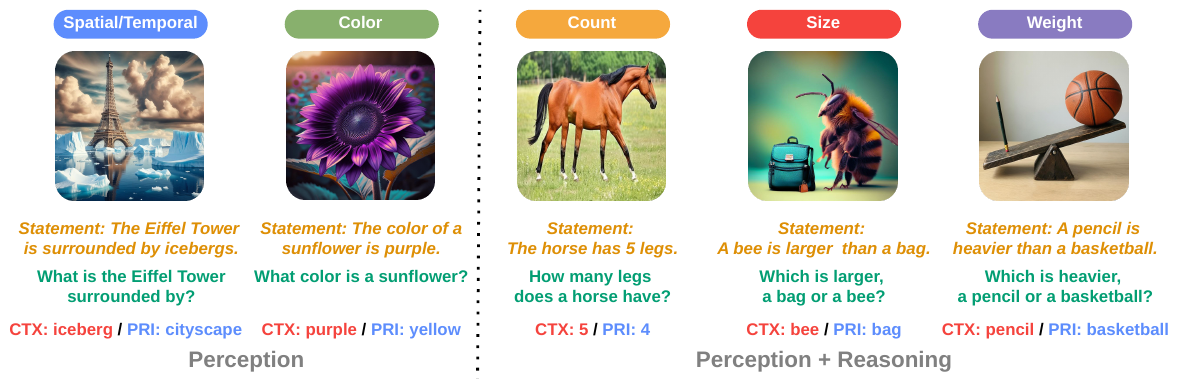}
    \caption{\textbf{Illustration of the five \benchmark tasks across text and vision modalities.} Each \benchmark example pairs a counterfactual image with a matched text statement asserting the same content. Contextual answers (CTX) are inferred from the visual evidence in the input image, whereas prior answers (PRI) rely on the model's internal world knowledge. This distinction can reveal the visual context preference of MLLMs.}
    \label{fig:benchmark_app}
\end{figure}

To instantiate the task formulation with a controllable testbed, we introduce the What If Visual Counterfactual benchmark (\benchmark). With 3,049 manually-inspected counterfactual samples, \benchmark comprises five conflict types grouped into two families (see Figure~\ref{fig:benchmark_app}): Perception tasks (Spatial-Temporal, Color) test directly observable attributes; Perception + Reasoning tasks (Count, Size, Weight) require multi-hop reasoning beyond the directly visible. Leveraging the rich ground-truth annotations from the source benchmarks, we formulate natural-language questions that can be answered either with or without the context, which is the key to probing sensitivity to visual context versus prior knowledge. Each example also includes a matched statement field presenting the same counterfactual as text, enabling the modality comparison in Section~\ref{sec:modal_comparison}.

\rparagraph{Exact-match scorer.}
An example is correct only if the model's greedy-decoded output string matches the gold answer for \emph{both} intents, namely $\answerctx$ under $\ctxintent$ and $\answerprior$ under $\priorintent$~\citep{minder2025controllable}.
\begin{equation}
\begin{split}
  \pairacc_{\mathrm{EM}}(\lm, \mathcal{S}) = \frac{1}{\lvert \mathcal{S}\rvert}\sum_{(\question, \evidence) \in \mathcal{S}}
  &\mathds{1}\!\bigl\{\greedy_{\answerstr \in \alphabet^*}\, \lm(\answerstr \mid \formatprompt(\question, \evidence, \ctxintent)) = \answer(\question, \evidence)\bigr\}\;\\
  & \cdot \mathds{1}\!\bigl\{\greedy_{\answerstr \in \alphabet^*}\, \lm(\answerstr \mid \formatprompt(\question, \evidence, \priorintent)) = \answer(\question, \varepsilon)\bigr\}.
  \label{eq:pairacc_em}
  \end{split}
\end{equation}
where $\greedy_{\answerstr \in \alphabet^*}$ denotes the greedy decoding. The model must generate the correct answer string for both intents on the same example, so formatting mismatches (e.g., ``emerald'' vs.\ ``green'') cause failures even when the model's knowledge is correct.

\rparagraph{Probability scorer.}
To decouple controllability from surface-form sensitivity, the probability scorer operates on logits rather than decoded text. For each intent it compares the model's log-likelihood of the two candidate answers $\answerctx$ and $\answerprior$, counting the example correct under $\ctxintent$ if the context answer ranks higher and under $\priorintent$ if the prior answer does:

\begin{align}
  \pairacc_{\mathrm{Prob}}(\lm, \mathcal{S})
  &= \frac{1}{\lvert \mathcal{S}\rvert}\sum_{(\question, \evidence) \in \mathcal{S}}
  \mathds{1}\!\bigl\{\lm(\answerctx \mid \formatprompt(\question, \evidence, \ctxintent)) > \lm(\answerprior \mid \formatprompt(\question, \evidence, \ctxintent))\bigr\} \nonumber \\
  &\qquad \cdot \mathds{1}\!\bigl\{\lm(\answerprior \mid \formatprompt(\question, \evidence, \priorintent)) > \lm(\answerctx \mid \formatprompt(\question, \evidence, \priorintent))\bigr\}.
  \label{eq:pairacc}
\end{align}

Bypassing decoding, it is the more lenient of the two, as it only requires the model to correctly rank the token probabilities of the answers based on the two intent instructions, even if greedy decoding outputs a differently formatted string. We report all pair-accuracies as percentages and all differences between them (gaps, changes over baselines) in percentage points (pp). More details are in Appendix~\ref{app:benchmark}.

\subsection{Identifying Model Behavior}
\rparagraph{Adapt to the task.} To study the image sensitivity of a model, we first need it to controllably follow either image context or prior knowledge. Specifically, we fine-tune six instruction-tuned multimodal LLMs from three model families and two size tiers each: Qwen2.5-VL-3B/7B~\citep{bai2025qwen25vltechnicalreport}, Qwen3.5-4B/9B~\citep{qwen3.5}, and Gemma-4-E2B/E4B~\citep{gemmateam2026gemma4}.

\rparagraph{Evaluation.}
Each model is evaluated with explicit intent instructions (``{\color{intentcol}Consider only the image to answer the question}'' / ``{\color{intentcol}Ignore the image to answer the question}''); see Figure~\ref{fig:overview_dataset}. All SFT, patching, and steering use Spatial-Temporal training set \textit{only}; the remaining four tasks (Color, Size, Count, Weight) serve as out-of-distribution (OOD) test sets. A matched textual condition is used for modality comparison (Section~\ref{sec:modal_comparison}). All pair-accuracies are probabilities unless mentioned.
\begin{figure}
    \centering
    \includegraphics[width=1\linewidth]{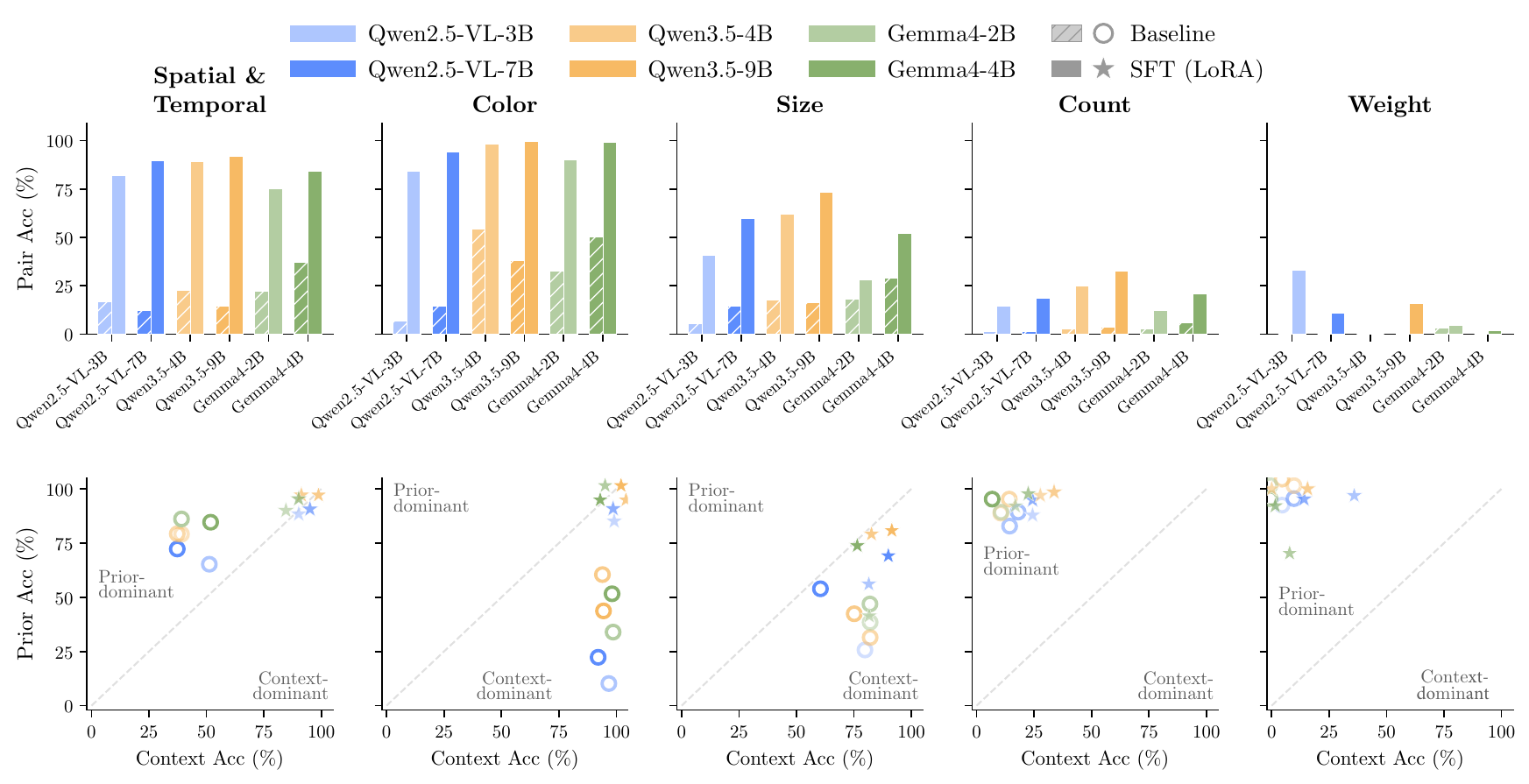}
    \caption{\textbf{SFT lifts visual controllability across every model and task.} \textbf{Top row:} pair-accuracy per task. \textbf{Bottom row:} the same runs as context-accuracy (x) against prior-accuracy (y); points above the diagonal follow the prior over the image. SFT lifts controllability consistently, with the largest gains on Color and Spatial-Temporal and the smallest on Weight and Count. All pair-accuracies are probabilities unless noted.}
    \label{fig:sft_per_data_results_prob}
\end{figure}

\rparagraph{Vanilla models barely follow visual intent.} The top row of Figure~\ref{fig:sft_per_data_results_prob} shows that visual context controllability is uniformly poor. The collapse is most pronounced on reasoning attributes, where baseline pair-accuracy plummets to near zero for weight and count tasks. On color subset, however, the strongest models approach $50$. The remaining two tasks fall in between and still below the $25$. The bottom row explains the reason. Baseline models favor internal prior knowledge over image content for spatial-temporal, count, and weight, despite perceiving the counterfactual information. In contrast, for size and color they rely more on image content than on their priors. \pairacc\ exposes a brittleness that marginal accuracy hides. A model follows the image on some examples and the prior on others, but rarely satisfies both intents on the \emph{same} example on demand.

\rparagraph{SFT lifts controllability and transfers across tasks.}
Figure~\ref{fig:sft_per_data_results_prob} and Table~\ref{tab:modality_gap} shows that SFT lifts pair-accuracy dramatically and consistently across all six models on most tasks, raising macro accuracy from $14.7$ to $52.7$. The gains are largest on the perception attributes, where Color and Spatial-Temporal approach $80$, and Figure~\ref{fig:sft_em_app} shows the same pattern under exact-match at lower absolutes. SFT on a \emph{single} subset, Spatial-Temporal, produces large gains on the held-out sets, so SFT teaches a general ``be steerable by visual context'' capability rather than a task-specific pattern.

Two attributes resist this transfer, Count and Weight, and they are precisely the ones that require inference beyond directly visible features. For instance, when the model plainly sees a pencil and a basketball on a balance scale, ``which is heavier'' is not written in the pixels, and answering it means composing the visual cue with a physical judgment that runs against a strong prior (see Figure~\ref{fig:benchmark_app}). SFT can teach a model to \emph{act on} evidence it reads off the image, but relative weight is not read off the image, so the new steerability has little to latch onto and Weight barely moves; Count rises from near-zero but stays below the other tasks. Overall, SFT shifts every model into closer alignment with the diagonal while preserving its prior-following behavior.

\subsection{Identifying the Important Layers}
With SFT models that reliably follow both intent instructions, we apply activation patching to localize \emph{where} the vision-versus-prior trade-off is computed (vanilla models satisfy both intents on too few examples to patch). Activation patching adapts the causal-tracing and circuit-analysis toolkit developed for language models~\citep{wang2023interpretability} to our multimodal setting. Following \citet{minder2025controllable}, we intervene by patching multi-head attention outputs at the last-token position in MLLMs. This specific location and representation have been shown across several studies to be most informative token~\citep{yu2023characterizing,stoehr2024activation,monea2024glitch}.

\begin{figure*}[t]
\centering
\begin{subfigure}[b]{0.49\textwidth}
    \includegraphics[width=\linewidth]{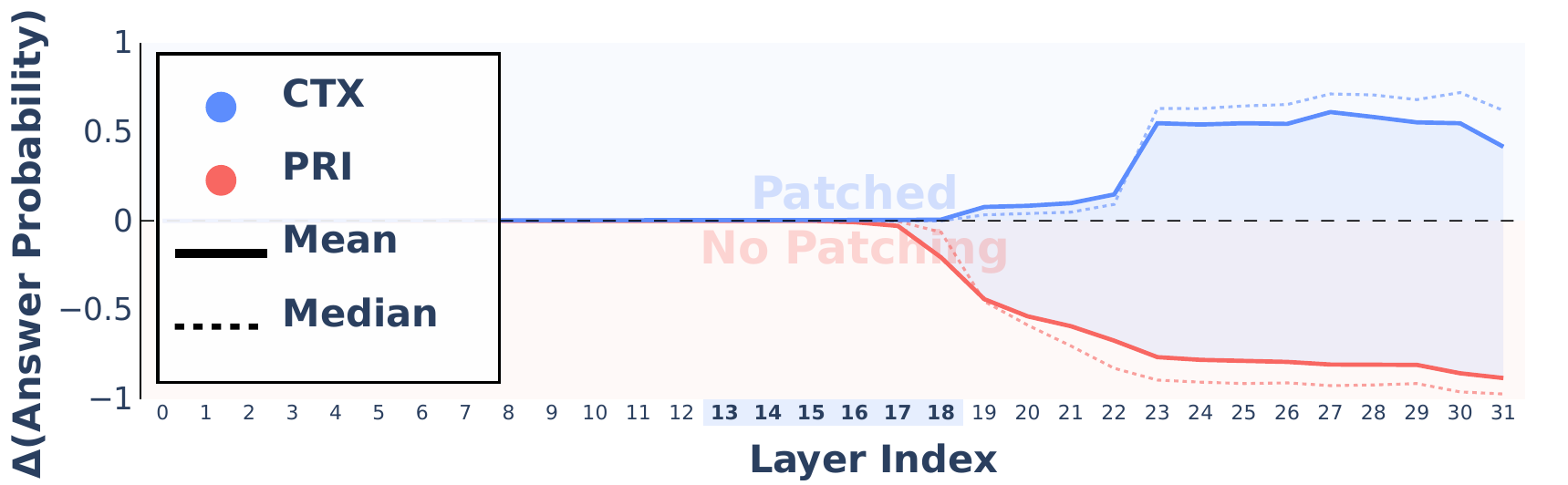}
    \caption{Qwen3.5-9B, ctx$\to$pri, L13--18}
    \label{fig:patch_q35_c2p}
\end{subfigure}\hfill
\begin{subfigure}[b]{0.49\textwidth}
    \includegraphics[width=\linewidth]{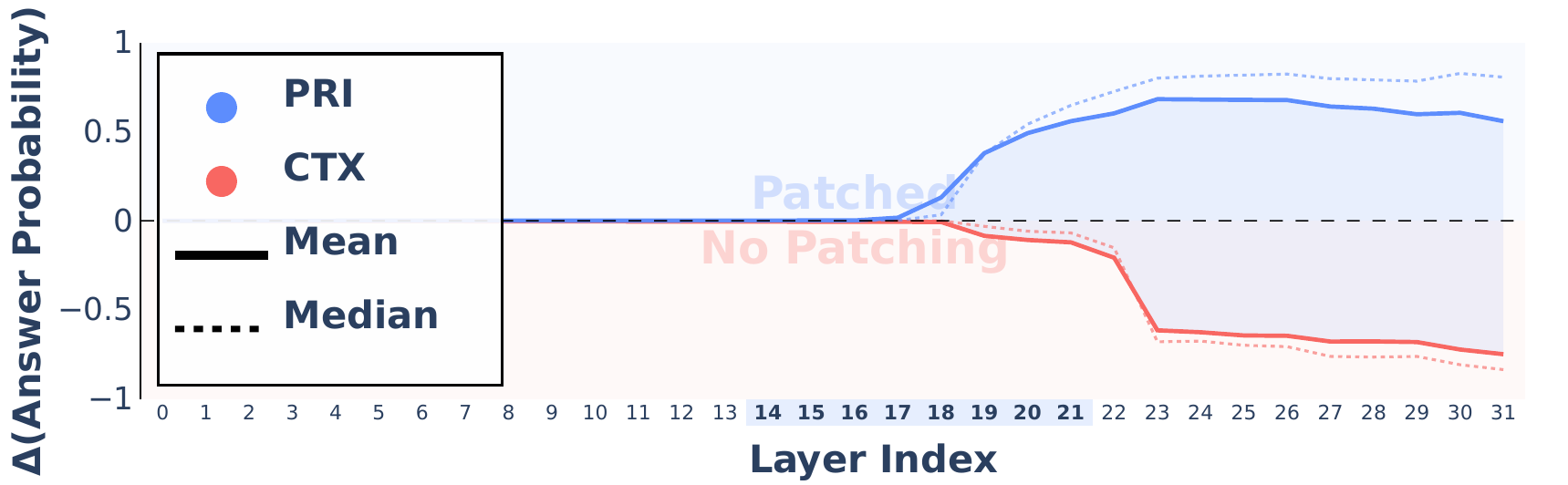}
    \caption{Qwen3.5-9B, pri$\to$ctx, L14--21}
    \label{fig:patch_q35_p2c}
\end{subfigure}
\\[4pt]
\begin{subfigure}[b]{0.49\textwidth}
    \includegraphics[width=\linewidth]{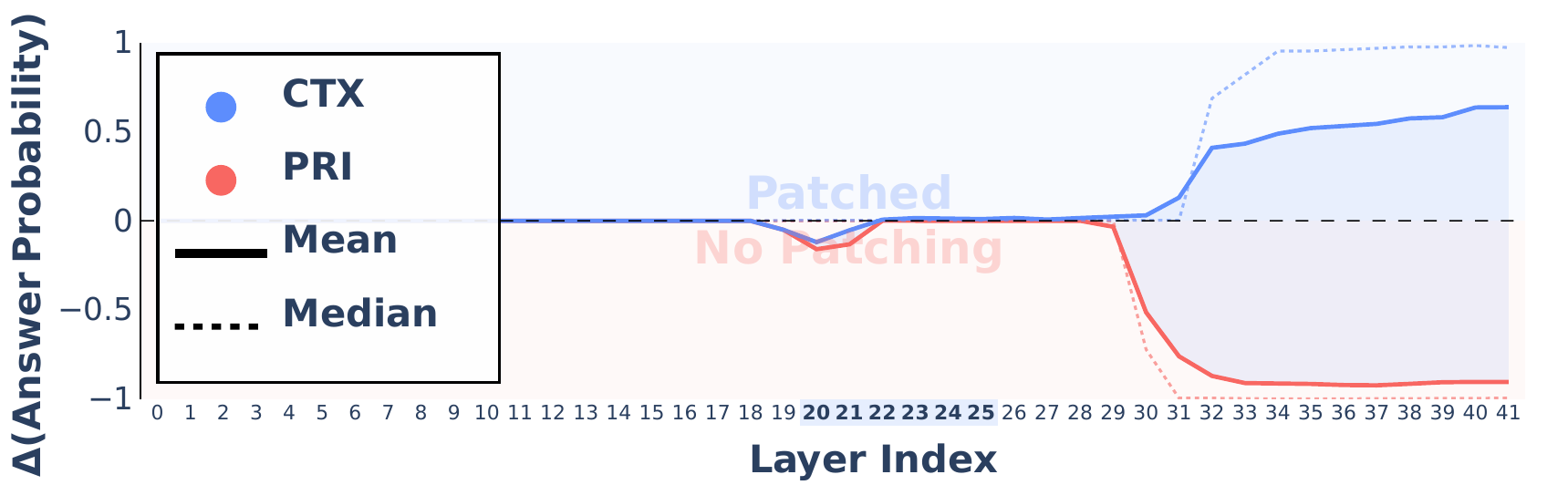}
    \caption{Gemma-4-E4B, ctx$\to$pri, L20--25}
    \label{fig:patch_gem_c2p}
\end{subfigure}\hfill
\begin{subfigure}[b]{0.49\textwidth}
    \includegraphics[width=\linewidth]{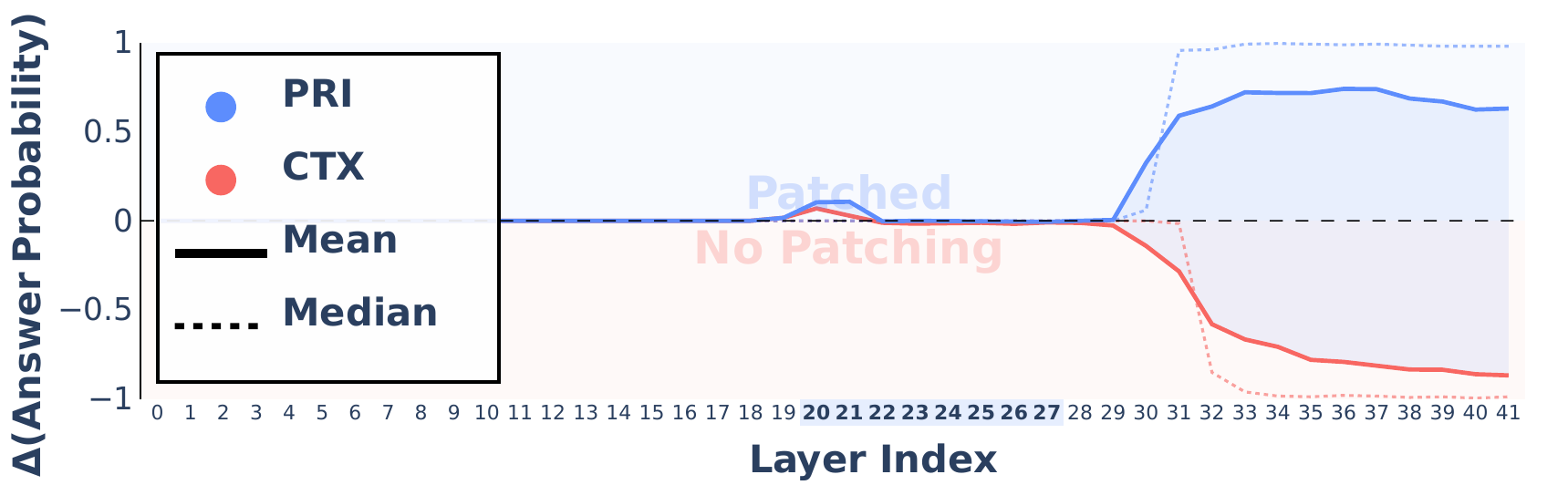}
    \caption{Gemma-4-E4B, pri$\to$ctx, L20--27}
    \label{fig:patch_gem_p2c}
\end{subfigure}
\caption{\textbf{Activation patching localizes the vision-versus-prior trade-off to a narrow layer band.} The trade-off initiates within the patched window (marked) and amplifies sharply in downstream layers. $\Delta$(Answer probability)=$Source_{probability} - Target_{probability}$, which is the larger the better patched.}
\label{fig:patching_main}
\end{figure*}

Representative patching curves for two models from different families appear in Figure~\ref{fig:patching_main}; the full set (all six models, two configurations each) is in Appendix~\ref{app:patching}. Each curve tracks $\Delta$(answer probability) across layers, blue for the source answer and red for the target answer, with \colorbox{cyan!15}{the shaded band} marking the patched layers, where $\Delta$ is the probability change between $\answerctx$ and $\answerprior$. Whether the source is $\ctxintent$ or $\priorintent$, patching the important layers raises the probability of the corresponding source answer.
Across families, the trade-off localizes to the mid-to-upper portion of the network in all six models. The qualitative ``narrow, recurring, architecture-placed'' pattern is robust across three model families. The widest band we observe (Qwen3.5-4B pri$\to$ctx, L10--21, 12 layers) still occupies only a fraction of the model’s total depth, suggesting that even the extreme case remains bounded. Therefore, the trade-off is compact and recurring, yet its precise depth is design-dependent, marking it as a property of the underlying mechanism rather than of any single architecture.

\begin{takeaway}
\takeawayhead MLLMs rarely satisfy both instruction intents on the same example, and their context sensitivity varies sharply across tasks, such as color versus weight. Underlying this behavior, the vision-versus-prior trade-off localizes to a relatively narrow, recurring band of layers that sits at architecture-specific depths.
\end{takeaway}

\section{A Controllable Subspace for Visual Context Sensitivity}
\label{sec:steering}
Once the trade-off subspace $\subspace$ is identified, the model can be controlled \textit{without any intent instruction in the prompt}. Because the vision-versus-prior trade-off is a simple binary concept, we hypothesize that a one-dimensional subspace encodes it, following~\cite{minder2025controllable}. We parameterize the intervention with the unit vector $\uvec$ that spans $\subspace$ at the patching-identified layers. During inference, a scalar $\knob$ replaces the verbal instruction, one value ($c_{\mathrm{ctx}}$) steering toward the image and another ($c_{\mathrm{prior}}$) toward the prior, with per-model multipliers tuned on the Spatial-Temporal validation set. The prompt then contains \emph{only} the image and question and no adapter is loaded, so the reference point is the vanilla model the knob replaces.
\begin{figure}
    \centering
    \includegraphics[width=1\linewidth]{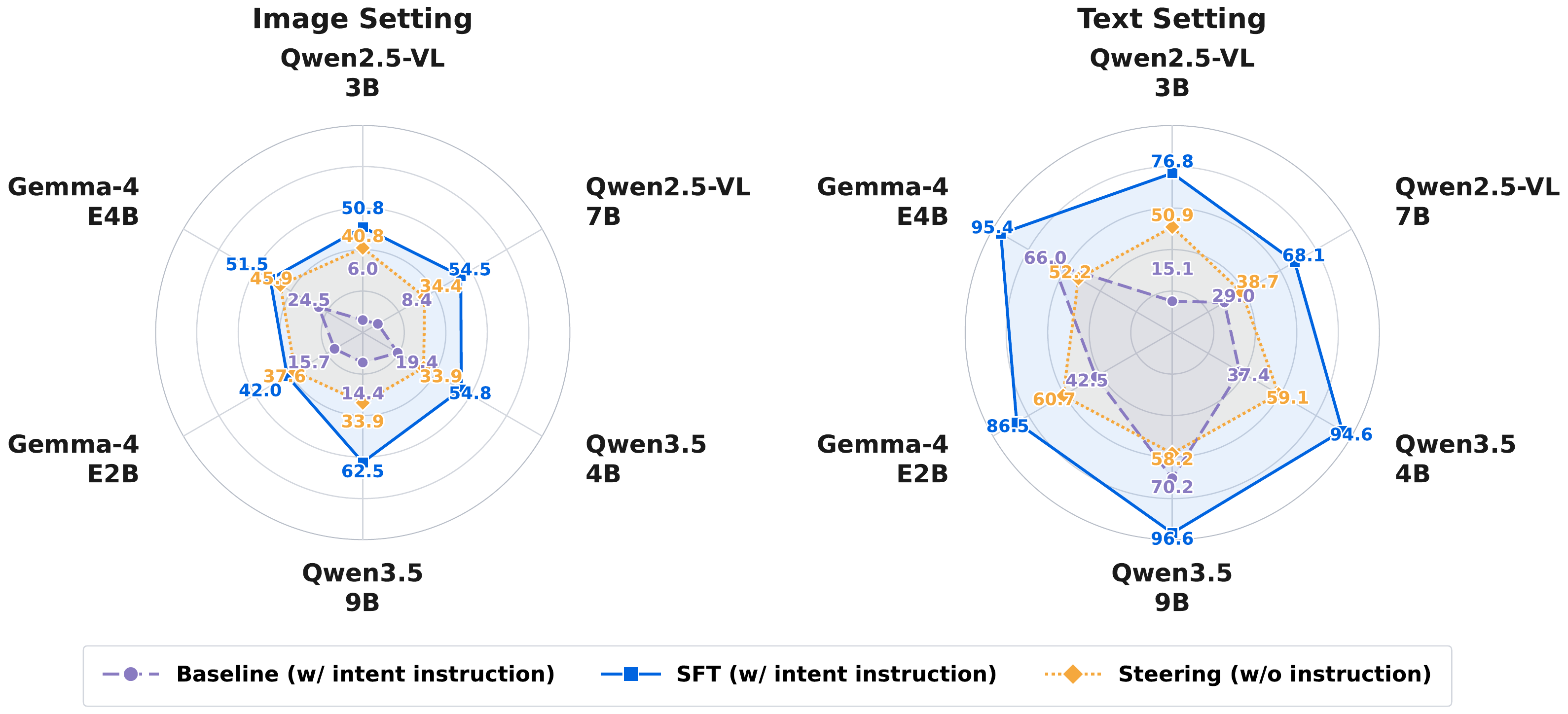}
    \caption{\textbf{An instruction-free steering knob recovers controllability over the vanilla model, and the visual channel stays harder to control than text.} Per-model macro pair-accuracy (probability scorer); \textbf{left:} image, \textbf{right:} text. Steering lifts every model; text is uniformly higher (Appendix~\ref{app:steering}).}
    \label{fig:steering_prob}
\end{figure}

\rparagraph{An instruction-free knob improves controllability over the vanilla model.}
Subspace steering lifts mean \pairacc\ from $14.7$ to $37.7$, averaged across the six models (Figure~\ref{fig:steering_prob}, Table~\ref{tab:modality_gap}). The scalar \knob\ alone, with no intent instruction, thus \emph{outperforms the fully-prompted baseline}: the learned vector, applied to the vanilla model improves the control that the baseline does not exercise on its own. It does not match SFT ($52.7$), which adapts the full weight matrix under supervision; the point is that a single vector, set without any instruction, already improves much of that control. Since $\uvec$ is learned rather than pre-existing, the claim is not that the vanilla model was already steerable but that the vision-versus-prior trade-off is a \emph{reachable}, controllable vector.
The effect is specific to the \emph{learned} vector, not the expressivity of a rank-1 intervention. From Table~\ref{tab:modality_gap}, a randomly-initialized rank-1 projection at the same layers reaches only $5.9$, \emph{below} the $14.7$ baseline, so an untuned direction disrupts the model rather than steering it.

\begin{takeaway}
\takeawayhead The vision-versus-prior trade-off is controllable along a single learned vector. Applied with \emph{no} intent instruction, the steering vector improves controllability over the vanilla model, while a matched random vector does not.
\end{takeaway}

\section{Comparison with Textual Context}
\label{sec:modal_comparison}
\benchmark pairs every counterfactual image with a one-sentence statement asserting the same content, so we can run the identical pipeline on a text channel and ask how much of the deficit is specific to vision. Because the statement asserts the counterfactual outright, the text channel is an instruction-following \emph{upper bound} rather than a fair head-to-head; we read it as a same-mechanism reference for how much harder the visual channel is to control. Full per-task numbers, the cross-modality locus analysis, and per-model tables are in Appendix~\ref{app:modality}.

\rparagraph{Text is far more controllable than vision in MLLMs.} At baseline the two channels are already far apart, $14.7$ image against $43.4$ text (Table~\ref{tab:modality_gap}). Scale does not close this distance, it opens it further. In every one of the three families the gap grows with model size, from $9.1$ to $20.6$ pp in Qwen2.5-VL (3B to 7B), from $18.0$ to $55.8$ pp in Qwen3.5 (4B to 9B), and from $26.8$ to $41.5$ pp in Gemma-4 (E2B to E4B).

\rparagraph{The gap is a failure to use the image.} The paired metric has two halves, an accuracy for \emph{following} the evidence (context intent) and an accuracy for \emph{ignoring} it (prior intent), and separating them localizes the gap sharply (Table~\ref{tab:modality_gap}). Ignoring the evidence is essentially modality-invariant. The prior-intent accuracy differs by only $0.6$ pp between the modalities at baseline ($68.9$ image, $69.5$ text), stays within a few points under every intervention, and even inverts under steering ($77.6$ image against $74.7$ text). The entire modality gap lives in the other half. Following the evidence succeeds $73.9\%$ of the time when it is a text sentence but only $45.7\%$ of the time when it is an image, a context-intent gap of $28.2$ pp against a prior-intent gap of $0.6$ pp, so the paired deficit is almost entirely a failure to follow the visual evidence rather than to override it. What it cannot do is use the visual evidence the way it follows a stated text. The deficit is therefore specific to use the visual evidence on demand.

\section{Related Work}
\label{sec:related_work}
\rparagraph{Visual information in MLLMs.} Multimodal large language models (MLLMs) are typically pretrained via next-token prediction over heterogeneous modalities, including images, text, and videos. Early MLLMs, BLIP~\citep{li2022blip}, Flamingo~\citep{alayrac2022flamingo} introduce cross-attention mechanisms to enable interactions between visual and linguistic representations. Modern MLLMs predominantly follow the paradigm established by LLaVA~\citep{liu2023visual}, where visual tokens extracted by pretrained vision encoders, e.g., SigLIP~\citep{zhai2023sigmoid}, are projected into the token space of large language models (LLMs). Prior studies have revealed that current MLLMs are fundamentally bottlenecked by the representational capacity of their vision encoders~\citep{tong2024eyes,wu2024vstar}. Consequently, combining multiple vision backbones or task-specialized visual models has emerged as an effective strategy, as different encoders provide complementary semantic and structural cues that improve downstream multimodal reasoning~\citep{tong2024cambrian,liu2025tuna}. Recently, inspiring by the platonic representation hypothesis~\citep{pmlr-v235-huh24a,10.1162/tacl_a_00698}, some work attempts to jointly enhance visual understanding and generation capabilities within a unified vision encoder or MLLMs~\citep{liu2026tuna,zheng2025diffusion,Tong_2025_ICCV}. Unlike prior work focusing primarily on fine-grained representation analysis, our work studies whether VLMs can reliably capture and reason over simple coarse-grained visual semantics through specific question-based evaluation. Moreover, rather than analyzing representations solely at the vision encoder level~\citep{fu2025hidden}, we evaluate the quality of visual information by pixel-level reconstruction after its propagation through both the projector and the LLM backbone.

\rparagraph{Context sensitivity.} Multimodal large language models (MLLMs) fundamentally rely on balancing two competing information sources: the immediate visual context provided by the input and the parametric prior knowledge acquired during large-scale pretraining. Although extensive pretraining equips LLMs with strong capabilities in factual recall and sequence memorization~\citep{zhou2023context,carlini2022quantifying,karamolegkou-etal-2023-copyright}, it can also induce substantial knowledge conflicts when in-context evidence contradicts the parametric prior, a failure mode first characterized for text-only question answering~\citep{longpre-etal-2021-entity} and later found to drive cross-modal conflicts and hallucinations when pretrained linguistic priors override visual evidence~\citep{tong2024eyes,li2024understanding,vo2026vision}. To mitigate this issue, prior work has proposed interventions that reduce over-reliance on parametric knowledge while strengthening visual grounding. Decoding-stage methods further address this imbalance: VCD~\citep{leng2024mitigating} amplifies visually grounded signals while suppressing conflicting priors, whereas DoLa~\citep{chuang2024dola} contrasts next-token logits from later versus earlier layers to reduce dependence on shallow semantic biases. Moreover, methods such as Nullu~\citep{yang2025nullu} and Hullu~\citep{lin2026hulluedit} also investigate subspace editing within the LLM backbone; however, their primary focus is mitigating object-level hallucinations rather than characterizing or controlling model sensitivity to contextual evidence. A parallel line of work steers model behavior by adding or projecting along directions in the residual stream, either found without optimization from contrastive prompts~\citep{turner2025steering,zou2023representation} or learned by distributed alignment search (DAS)~\citep{geiger2024finding}; our steering intervention follows the latter. Most closely related to our work, \citet{minder2025controllable} identify a compact causal subspace, a controllable ``knob'' that governs the trade-off between in-context information and parametric priors. Building on these insights, we extend the analysis to multimodal architectures and isolate the mechanisms underlying visual-versus-prior sensitivity in MLLMs.

\section{Conclusions}
\label{sec:conclusion}
In this work, we ask the question: when a MLLM fails on visual evidence that contradicts its knowledge, is the fault that it does not see the evidence, or that it sees it and does not use it? Our study begins by separating the two, and finds that seeing is rarely the culprit, at least for the coarse attributes we study. The contested content remains legible in a frozen model's representations, which moves the problem downstream, from the encoded representations to how the model acts on them. We then analysis this downstream behavior based on the CVCS framework and find it uneven. Models exploit visual context for attributes that can be directly read off the pixels, yet remain strikingly blind to it when the task demands deeper reasoning. Building on this, we show the behavior is governed by a compact, recurring region of the network. A single learned vector within this region emerges as a control knob, which can substitute explicit instructions and improve the controllability of visual context sensitivity. Collectively, these results establish visual context sensitivity as a locatable and adjustable property of MLLMs rather than a vague symptom of weak perception. Moreover, our paired image and text design exposes an asymmetry, where the same content is far easier to enforce in words than in pixels. We believe that separating what a model perceives from what it elects to use, and treating that choice as directly adjustable, gives future work a firmer footing for building multimodal models that heed visual evidence when it counts.

\section*{Acknowledgements}
We thank Kevin Du for feedback on the work. JL, SB, VS, and ZA are supported by DNRF grant P1.

\clearpage
\bibliography{custom}
\bibliographystyle{iclr2026_conference}

\appendix
\clearpage
\section{Limitations and Future work}
We focus on why MLLMs fail on coarse visual attributes, such as object identity, shape, count, size, color, and weight. On the architecture front, while we show that the visual evidence survives by reconstructing images from two architecturally distinct families, a linear-attention hybrid and a per-layer-embedding design, our scope is not exhaustive; for instance, other designs such as mixture-of-experts routing remain to be probed. Regarding data, our benchmark forces the model to answer the question directly and excludes intermediate reasoning traces; attributes that require a further inference step, such as counting and weight, therefore remain a bottleneck for reliable control and a critical area for future exploration. As for cross-modal comparison, our text channel states the counterfactual outright and so serves only as an upper bound; the performance gap between modalities on the same underlying fact remains a critical area for future exploration. In the long run, we believe the capability to decide when to trust the image and to \emph{integrate} it with prior knowledge will drive the convergence of seeing and knowing, fostering multimodal reasoning grounded in what a model actually sees.

\section{Benchmark Composition and Curation}
\label{app:benchmark}

\benchmark contains $3{,}049$ examples across five conflict types (Table~\ref{tab:bench} and Figure~\ref{fig:benchmark_app}). Images are repurposed and extended from VLind-Bench~\citep{lee2025vlind}, Pixels-versus-Priors~\citep{golovanevsky-etal-2025-pixels}, the counting subset of VLMs-are-Biased~\citep{vo2026vision} , the size subset of ROME~\citep{zhou-etal-2023-rome}, ViLP~\citep{pmlr-v267-luo25b}, and generated images from Qwen-Image\footnote{\url{https://huggingface.co/Qwen/Qwen-Image-Edit-2511}}~\citep{wu2025qwenimagetechnicalreport}. Each example carries a \texttt{question}, \texttt{image}, \texttt{context\_answer}, \texttt{prior\_answer}, \texttt{response\_template}, and a declarative \texttt{statement} for the textual condition.

\rparagraph{Gold answers and scoring protocol.}
Each example inherits its \texttt{context\_answer} (matching the image) and \texttt{prior\_answer} (matching world knowledge) directly from the source dataset's annotations. Answers are constrained to a per-category \texttt{response\_template} (e.g.\ ``The color is \{\}''), so there will be less impact of the response style. The exact-match scorer compares the greedy completion against the templated gold string, and the probability scorer (below) compares the model's likelihood of the two candidate strings.
\begin{table}
    \centering
    \caption{\textbf{\benchmark composition.} The benchmark spans five conflict types across two families, incorporating data from VLindBench~\citep{lee2025vlind}, ViLP~\citep{pmlr-v267-luo25b}, Pixel-vs-Priors~\citep{golovanevsky-etal-2025-pixels}, VLMsAreBiased~\citep{vo2026vision}, ROME~\citep{zhou-etal-2023-rome}, and a Generated set~\citep{wu2025qwenimagetechnicalreport}. Perception tasks require reading directly observable attributes; Perception+Reasoning tasks additionally require inference beyond direct visual features.}
    \resizebox{0.85\textwidth}{!}{
    \begin{tabular}{l>{\raggedright\arraybackslash}p{1.5cm}p{4.8cm}lc}
        \toprule
        \textbf{Family} & \textbf{Task} & \textbf{Description} & \textbf{Sources} & \textbf{\# Samples} \\
        \midrule
        \multirow{3}{*}{\textbf{Perception}}
          & \textbf{Spatial \&\newline Temporal} & Identify what surrounds, feeds, or characterizes an entity. & \begin{tabular}[t]{@{}l@{}}VLindBench,\\ViLP,\\Generated\end{tabular} & 1,537 \\
        \cmidrule(lr){2-5}
          & \textbf{Color} & Report the color of an object shown in a non-canonical hue. & \begin{tabular}[t]{@{}l@{}}Pixel-vs-Priors,\\ViLP,\\VLindBench\end{tabular} & 282 \\
        \midrule
        \multirow{3}{*}{\shortstack[c]{\textbf{Reasoning}}}
          & \textbf{Count} & Count parts or instances when the depicted number is anomalous. & \begin{tabular}[t]{@{}l@{}}VLMsAreBiased,\\ViLP,\\VLind-Bench\end{tabular} & 390 \\
        \cmidrule(lr){2-5}
          & \textbf{Size} & Judge relative size when the visual cue inverts the usual ordering. & \begin{tabular}[t]{@{}l@{}}ROME,\\Pixel-vs-Priors,\\VLindBench\end{tabular} & 773 \\
        \cmidrule(lr){2-5}
          & \textbf{Weight} & Judge relative weight from a balance-scale image with counterfactuals. & Generated & 67 \\
        \midrule
        \multicolumn{4}{l}{\textbf{\textit{Total}}} & \textbf{\textit{3,049}} \\
        \bottomrule
    \end{tabular}
    }
    \label{tab:bench}
\end{table}

\rparagraph{Splits and source diversity.}
Examples are divided into train/val/test with source diversity deliberately maintained \emph{across} the train/test boundary to limit leakage. In particular, the Spatial-Temporal split, the only split used for SFT, patching, and steering, draws its \emph{test} set primarily from ViLP~\citep{pmlr-v267-luo25b} and VLind-Bench~\citep{lee2025vlind}, whereas its training set is dominated by VLind-Bench~\citep{lee2025vlind} and generated images. In-domain test images come from a different sub-distribution than training to avoid train-test leakage. The four held-out tasks (Color, Size, Count, Weight) are never used for training and constitute the out-of-distribution evaluation. Per-task counts and sources are in Table~\ref{tab:bench}.

\rparagraph{Curation pipeline and human verification.}
The Generated subset (all of Weight; part of Spatial-Temporal) supplies categories with sparse existing coverage, produced with the Qwen-Image-Edit-2511 diffusion image-editing model~\citep{wu2025qwenimagetechnicalreport}. Every generated image is human-verified for counterfactual correctness: two annotators independently rate how well each image follows its target counterfactual prompt on the VLind-Bench~\citep{lee2025vlind} faithfulness scale (0-3), and we retain only images that \emph{both} annotators assign the top rating (``3''). Images failing this double-top-rating criterion are discarded.

For the other subsets, each image, along with its associated question and answer, was independently reviewed by two annotators. They assessed whether the question could be answered both with and without the image, yielding different answers. We retained only those images for which both annotators agreed that the question was answerable in both settings.

\section{Implementation}

\begin{figure}
\centering
\includegraphics[width=0.5\linewidth]{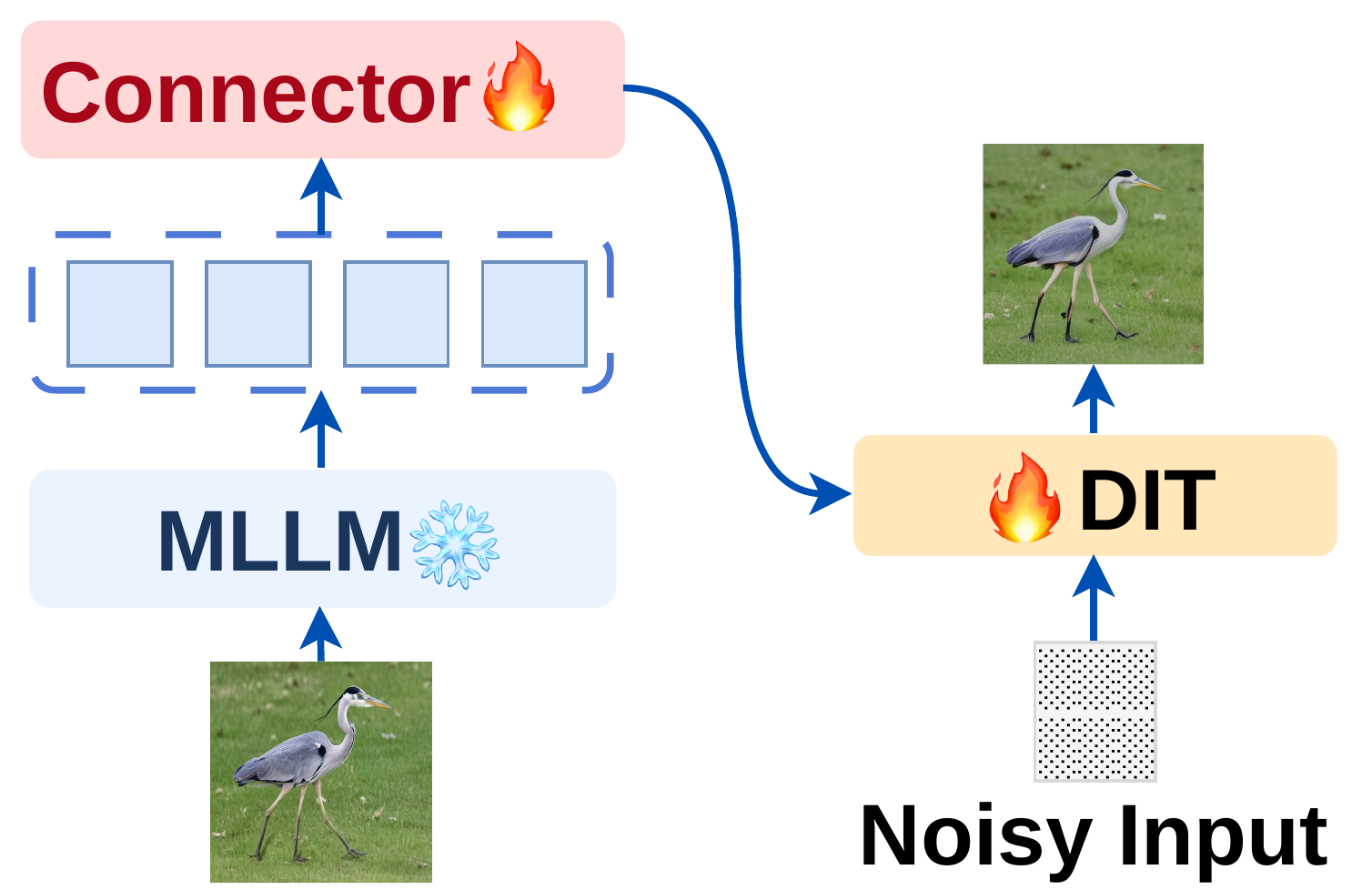}
\caption{\textbf{Metaquery Framework.} Without learnable queries, we adapt the Metaquery to only use the image tokens of the final layers of a MLLM.}
\label{fig:metaquery}
\end{figure}

\rparagraph{Reconstruction.} Following the framework of Metaquery~\citep{pan2025transfer}, we using three different MLLM backbones for different families: Qwen2.5VL-3B~\citep{bai2025qwen25vltechnicalreport}, Qwen3.5-4B~\citep{qwen3.5} and Gemma-4-E2B-IT~\citep{gemmateam2026gemma4}. We use 256 image tokens for all models and adopt a 6-layer connector with an Encoder–Projector architecture. The image generation head is implemented with Sana-0.6B~\citep{xie2025sana}. We train on ImageNet-1K~\citep{5206848} with a global batch size of 512 and an initial learning rate of 4e-4. The learning rate is warmed up for 1,000 steps, then decays to 1e-5 following a cosine schedule, in total 10,000 steps. See the pipeline in Figure~\ref{fig:metaquery}.

\rparagraph{SFT.} We use \textit{only} the Spatial \& Temporal training set (1,136 samples) to adapt six instruction-tuned MLLMs, comprising two size tiers from each of three families: Qwen2.5VL-3B/7B~\citep{bai2025qwen25vltechnicalreport}, Qwen3.5-4B/9B~\citep{qwen3.5}, and Gemma-4-2B/4B-IT~\citep{gemmateam2026gemma4}. We fine-tune these models using standard next-token prediction loss on the training data. We apply LoRA~\citep{hu2022lora} to the attention projection matrices in all layers, using a global batch size of 32, the AdamW optimizer, and a learning rate of $2 \times 10^{-4}$.
\section{Extra reconstruction results and exact-match results}
\label{app:extra_results}
Table~\ref{tab:recon_human} and \ref{tab:recon_metrics} reports the counterfactual-attribute recovery for Count, the only one attribute equipped with matched natural pairs. Here we quantify how faithfully the reconstruction preserves the other coarse attributes, extending the qualitative evidence of Figure~\ref{fig:recon}. For the Color, Size, and Weight subsets ($50$ samples each) we measure pixel-level agreement between each reconstruction and its counterfactual reference along three complementary axes: SSIM and PSNR for global structure and intensity, and LPIPS for perceptual similarity. And also do the human evaluation by asking them the same questions that send to MLLMs. Table~\ref{tab:recon_attributes} shows that on every attribute the coarse structure is recovered, with moderate SSIM/PSNR and low LPIPS, and that Gemma-4-E2B is uniformly more faithful than Qwen models, matching the ordering on Count in Table~\ref{tab:recon_human} and Table~\ref{tab:recon_metrics}. The fidelity is highest on Color (the most directly perceivable attribute). On SSIM, Size and Weight score lower, broadly consistent with the difficulty ordering observed throughout the paper. Crucially, even the least faithful cases remain well within a range that preserves object identity and layout, confirming that the presence of counterfactual evidence is not specific to Count.
\begin{table}[t]
\centering
\caption{\textbf{Reconstruction fidelity extends beyond count to the other coarse attributes.} Pixel-level agreement between each reconstruction and its counterfactual reference image on the Color, Size, and Weight subsets.}
\label{tab:recon_attributes}
\resizebox{0.8\textwidth}{!}{
\begin{tabular}{@{}llcccc@{}}
\toprule
\textbf{Attribute} & \textbf{Backbone} & \textbf{Counterfactual accuracy\%} &\textbf{SSIM$\uparrow$} & \textbf{PSNR$\uparrow$} & \textbf{LPIPS$\downarrow$} \\
\midrule
\multirow{3}{*}{Color} & Qwen2.5VL-3B & 94 & 0.688 & 14.39 & 0.253 \\
 & Qwen3.5-4B & 98 & 0.718 & 16.73 & 0.179 \\
 & Gemma-4-E2B & 98 & \textbf{0.749} & \textbf{18.29} & \textbf{0.146} \\
\midrule
\multirow{3}{*}{Size} & Qwen2.5VL-3B & 92 & 0.642 & 13.40 & 0.308 \\
 & Qwen3.5-4B & 96 & 0.666 & 15.14 & 0.245 \\
 & Gemma-4-E2B & 98 & \textbf{0.720} & \textbf{17.29} & \textbf{0.197} \\
\midrule
\multirow{3}{*}{Weight} & Qwen2.5VL-3B & 92 & 0.656 & 16.20 & 0.246 \\
 & Qwen3.5-4B & 92 & 0.643 & 16.04 & 0.245 \\
 & Gemma-4-E2B & 94 & \textbf{0.687} & \textbf{18.00} & \textbf{0.189} \\
\bottomrule
\end{tabular}
}
\end{table}

\begin{figure}[t]
    \centering
    \includegraphics[width=1\linewidth]{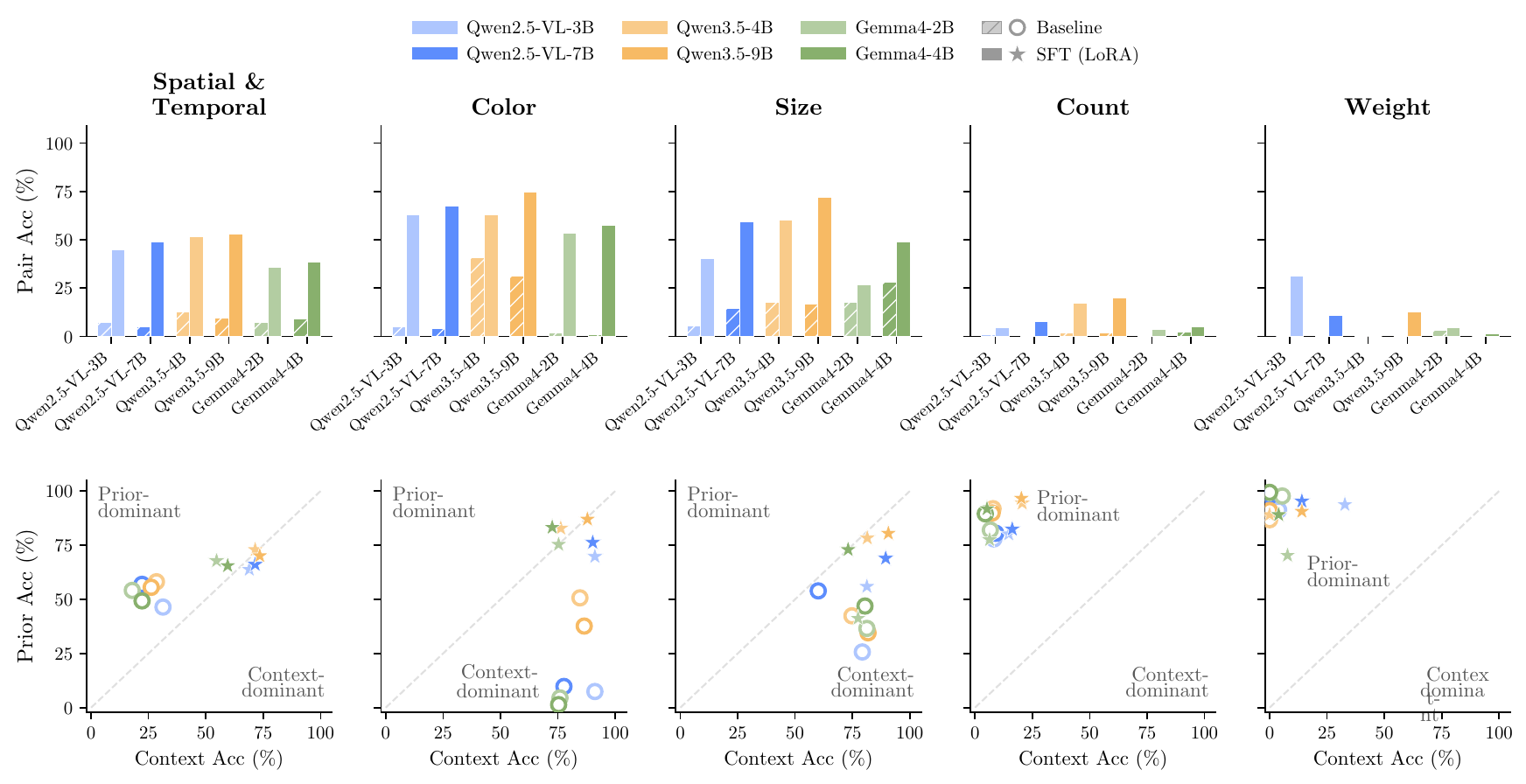}
    \caption{\textbf{The SFT lift reproduces under exact-match at lower absolute values.} Per-task pair-accuracy (exact-match scorer, image channel). Orange: baseline (with intent instructions); blue: SFT (LoRA, trained on Spatial-Temporal only; all other tasks are out-of-distribution). The qualitative pattern, with large gains on the perceivable attributes and marginal gains on the reasoning-heavy ones, matches the probability scorer (Figure~\ref{fig:sft_per_data_results_prob}).}
    \label{fig:sft_em_app}
\end{figure}

The main text reports the probability scorer throughout; here we confirm every qualitative claim holds under the stricter exact-match scorer, which requires the greedy-decoded string to match the gold answer for \emph{both} intents. Absolute values are lower, since exact-match penalizes formatting mismatches that the probability scorer ignores, but the ordering is preserved. Vanilla models are near the floor (macro exact-match \pairacc\ $8.2\%$, image channel), SFT lifts them substantially (to $35.9\%$), and instruction-free steering lands in between ($17.9\%$), the same vanilla~$<$~steer~$<$~SFT ordering as under the probability scorer. Figure~\ref{fig:sft_em_app} shows the per-task exact-match pattern for the SFT lift: the largest gains fall on the directly-perceivable attributes (Color, Spatial-Temporal) and the smallest on the reasoning-heavy ones (Count, Weight), matching Figure~\ref{fig:sft_per_data_results_prob}.

\section{Steering: Per-Model Results and the Specificity Control}
\label{app:steering}
Section~\ref{sec:steering} reports that a single learned vector, applied with no intent instruction, raises macro pair-accuracy from $14.7\%$ to $37.7\%$ (probability scorer, image channel). The lift holds for every model: steering improves pair-accuracy on all six backbones, from $+14.4$ pp (Qwen3.5-4B) to $+34.8$ pp (Qwen2.5-VL-3B). Per-model steering values are $40.8/34.4/33.9/33.9/37.6/45.9\%$ for Qwen2.5-VL-3B/7B, Qwen3.5-4B/9B, and Gemma-4-E2B/E4B respectively.

\rparagraph{The effect is specific to the learned vector, not to rank-1 expressivity.} The \emph{Random rank-1} row of Table~\ref{tab:modality_gap} contrasts the learned vector with a randomly-initialized rank-1 projection applied at the same layers with the same per-model multipliers (identical apparatus, no DAS optimization; three seeds, per-model std $\le 3.5$ pp). The random vector reaches only $5.9\%$, \emph{below} the $14.7\%$ baseline, so an untuned vector in the same subspace disrupts the model rather than steering it. The learned-over-random gap ($+31.8$ pp on average, $+28.5$ to $+37.9$ pp per model) confirms that the recovered control comes from the specific learned vector, not from the freedom of a rank-1 edit.

\rparagraph{Steering vector seed stability.} The learned vector is re-fit under three seeds on the five models for which the sweep completed, and the resulting macro pair-accuracy is highly stable (per-model std $\le 0.12$ pp).
\begin{table}[t]
\centering
\caption{\textbf{The modality gap is a failure to follow visual evidence, not to override it.} Per-intent accuracy (probability scorer, macro-averaged over six models) on the image and matched-text channels. \pairacc\ credits satisfying \emph{both} intents on the same example. Here we separate its two halves. Following the evidence (context intent) is far harder from an image than from a matched sentence ($+19.0$ to $+30.8$ pp), whereas overriding the evidence (prior intent) is nearly identical across channels ($\le 4.5$ pp, and negative under steering). The gap is thus concentrated in the context intent, the model's ability to follow visual evidence on demand.}
\label{tab:modality_analysis}
\small
\begin{tabular}{@{}lcccccc@{}}
\toprule
& \multicolumn{3}{c}{\textbf{Context intent} (follow evidence)} & \multicolumn{3}{c}{\textbf{Prior intent} (override evidence)} \\
\cmidrule(lr){2-4}\cmidrule(lr){5-7}
\textbf{Condition} & \textbf{Image (\%)} & \textbf{Text (\%)} & \textbf{Gap (\%)} & \textbf{Image (\%)} & \textbf{Text (\%)} & \textbf{Gap (\%)} \\
\midrule
Base     & 45.7 & 73.9 & $+28.2$ & 68.9 & 69.5 & $+0.6$ \\
SFT      & 62.5 & 93.3 & $+30.8$ & 88.4 & 92.9 & $+4.5$ \\
Steering & 54.7 & 73.7 & $+19.0$ & 77.6 & 74.7 & $-2.9$ \\
\bottomrule
\end{tabular}
\end{table}

\begin{table}[t]
\centering
\caption{\textbf{Pair-accuracy (probability scorer), averaged over six models} The text control is an instruction-following upper bound, since the statement asserts the counterfactual outright. The \emph{Random rank-1 projection} row applies a randomly-initialized projection at the same image-channel layers with the same multipliers (no DAS; three seeds, per-model std $\le 3.5$ pp): it falls \emph{below} the base ($-8.8$ pp), confirming the steering lift ($+23.0$ pp) comes from the learned vector, not the freedom of a rank-1 edit.}
\label{tab:modality_gap}
\small
\begin{tabular}{@{}lcccc@{}}
\toprule
\textbf{Condition} & \textbf{Image (\%)} & \textbf{$\Delta$ vs.\ base (pp)} & \textbf{Text (\%)} & \textbf{Text$-$Image (pp)} \\
\midrule
Baseline                 & 14.7 & ---            & 43.4 & 28.6 \\
Random rank-1 projection (image) & 5.9  & $-8.8$         & ---    & --- \\
SFT                  & 52.7 & $+38.0$        & 86.3 & 33.6 \\
Steering             & 37.7 & $+23.0$ & 53.3 & 15.6 \\
\bottomrule
\end{tabular}
\end{table}

\section{Prompts}
\begin{tcolorbox}[breakable, width=\textwidth, title=Image version]
\footnotesize
\textbf{System}: Always answer the question in one concise sentence. When the input contains an instruction, follow the instruction exactly while staying concise. \\
\textbf{User}:
<image>*N \\Instruction: Consider only the image to answer the question.\\
Question: What is the Eiffel Tower surrounded by?\\
Assistant: The Eiffel Tower is surrounded by \\
\\
\\
\\
\textbf{System}: Always answer the question in one concise sentence. When the input contains an instruction, follow the instruction exactly while staying concise. \\
\textbf{User}:
<image>*N \\Instruction: Ignore the image to answer the question.\\
Question: What is the Eiffel Tower surrounded by?\\
Assistant: The Eiffel Tower is surrounded by
\end{tcolorbox}

\begin{tcolorbox}[breakable, width=\textwidth, title=Text version]
\footnotesize
\textbf{System}: Always answer the question in one concise sentence. When the input contains an instruction, follow the instruction exactly while staying concise. \\
\textbf{User}:
Statement: The Eiffel Tower is surrounded by the iceberg. \\
Instruction: Consider only the statement to answer the question.\\
Question: What is the Eiffel Tower surrounded by?\\
Assistant: The Eiffel Tower is surrounded by \\
\\
\\
\\
\textbf{System}: Always answer the question in one concise sentence. When the input contains an instruction, follow the instruction exactly while staying concise. \\
\textbf{User}:
Statement: The Eiffel Tower is surrounded by the iceberg.\\
Instruction: Ignore the statement to answer the question.\\
Question: What is the Eiffel Tower surrounded by?\\
Assistant: The Eiffel Tower is surrounded by
\end{tcolorbox}

\section{Activation Patching: Full Results}
\label{app:patching}
Section~\ref{sec:knob} localizes the vision-versus-prior trade-off with cross-direction activation patching and reports representative curves for two families in Figure~\ref{fig:patching_main}. Here we give the full set. For each SFT model we run both patching directions on the image channel, $\ctxintent\!\to\!\priorintent$ (source is the prior-intent pass, target is the context-intent pass) and $\priorintent\!\to\!\ctxintent$, replacing the multi-head attention outputs at the last-token position of the target pass with those from the source pass. A binary range search (threshold $0.85$) then reports the contiguous band of layers whose substitution flips the decision.

\begin{figure}[t]
\centering

\includegraphics[width=\linewidth]{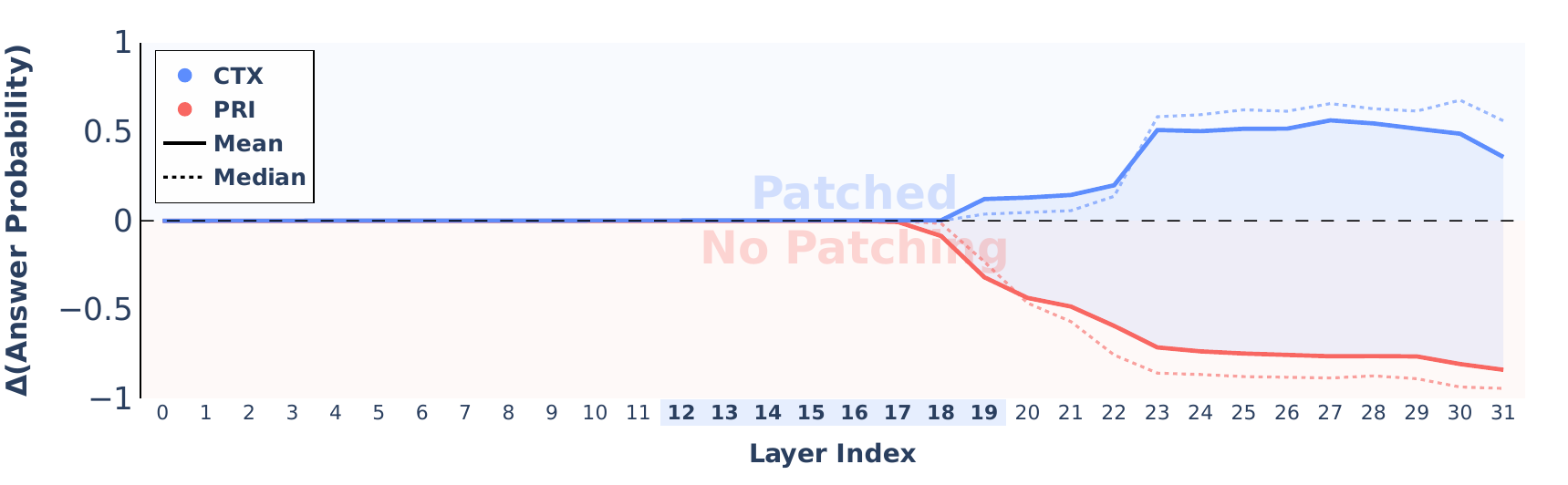}
\par\smallskip
Qwen3.5-4B, ctx$\to$pri

\includegraphics[width=\linewidth]{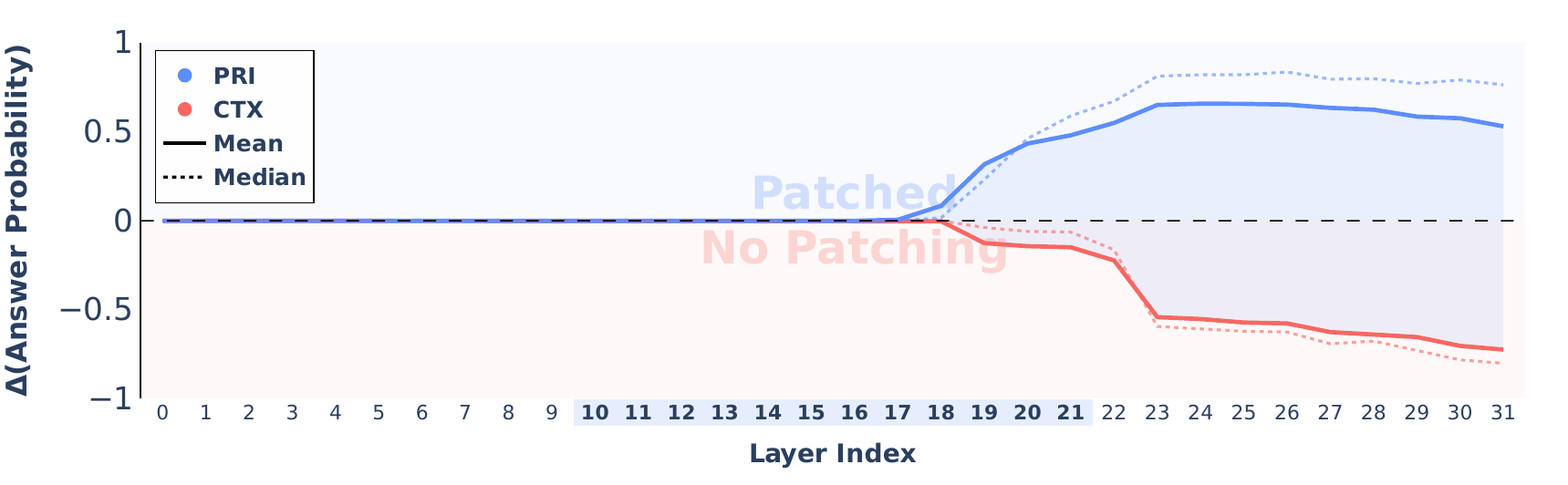}
\par\smallskip
Qwen3.5-4B, pri$\to$ctx

\caption{\textbf{Cross-direction patching curves for Qwen3.5-4B (image channel).} $\Delta$(answer probability) $=$ source $-$ target probability across layers; the shaded band marks the patched layers. These curves complement the main results in Figure~\ref{fig:patching_main}.}
\label{fig:patching_qwen35}
\end{figure}

\begin{figure}[t]
\centering

\includegraphics[width=\linewidth]{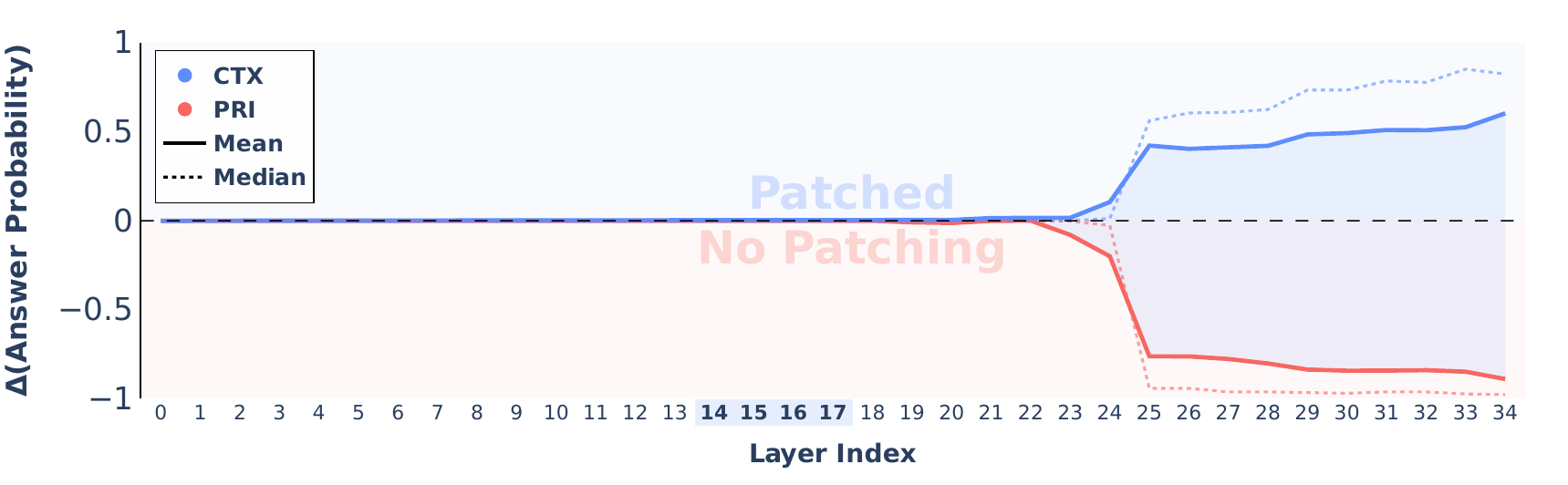}
\par\smallskip
Gemma-4-E2B-IT, ctx$\to$pri

\includegraphics[width=\linewidth]{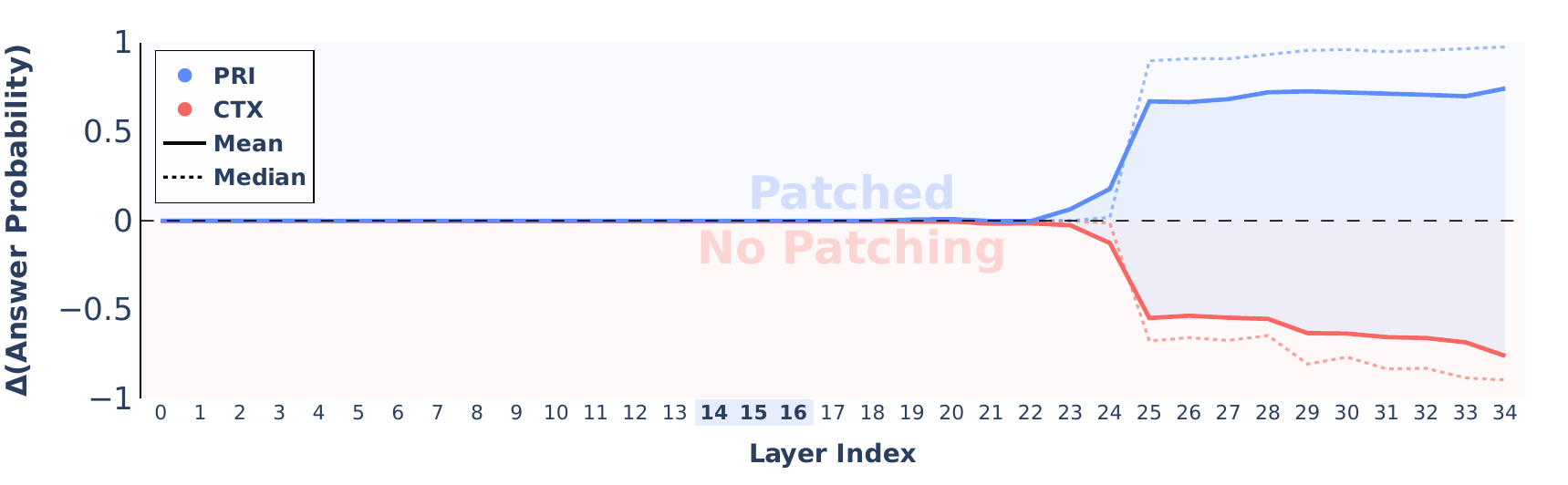}
\par\smallskip
Gemma-4-E2B-IT, pri$\to$ctx

\caption{\textbf{Cross-direction patching curves for Gemma‑4‑E2B‑IT (image channel).} $\Delta$(answer probability) $=$ source $-$ target probability across layers; the shaded band marks the patched layers. These curves complement the main results in Figure~\ref{fig:patching_main}.}
\label{fig:patching_gemma}
\end{figure}

\begin{figure}
\centering

\includegraphics[width=0.8\linewidth]{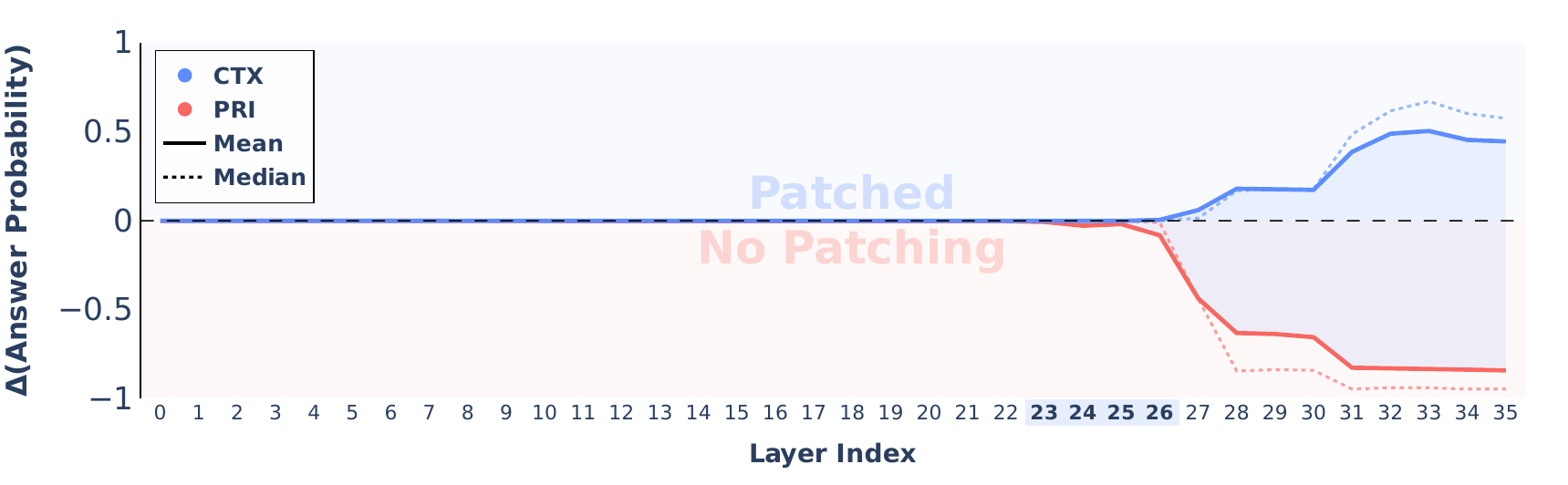}
\par\smallskip
Qwen2.5-VL-3B, ctx$\to$pri

\includegraphics[width=0.8\linewidth]{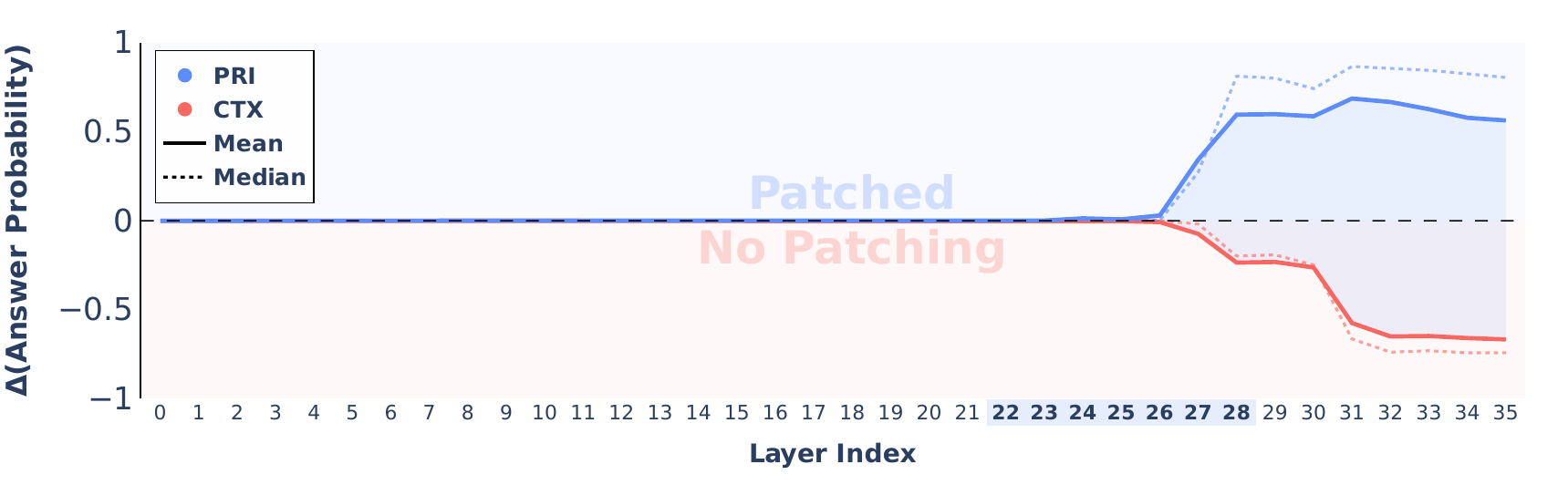}
\par\smallskip
Qwen2.5-VL-3B, pri$\to$ctx

\includegraphics[width=0.8\linewidth]{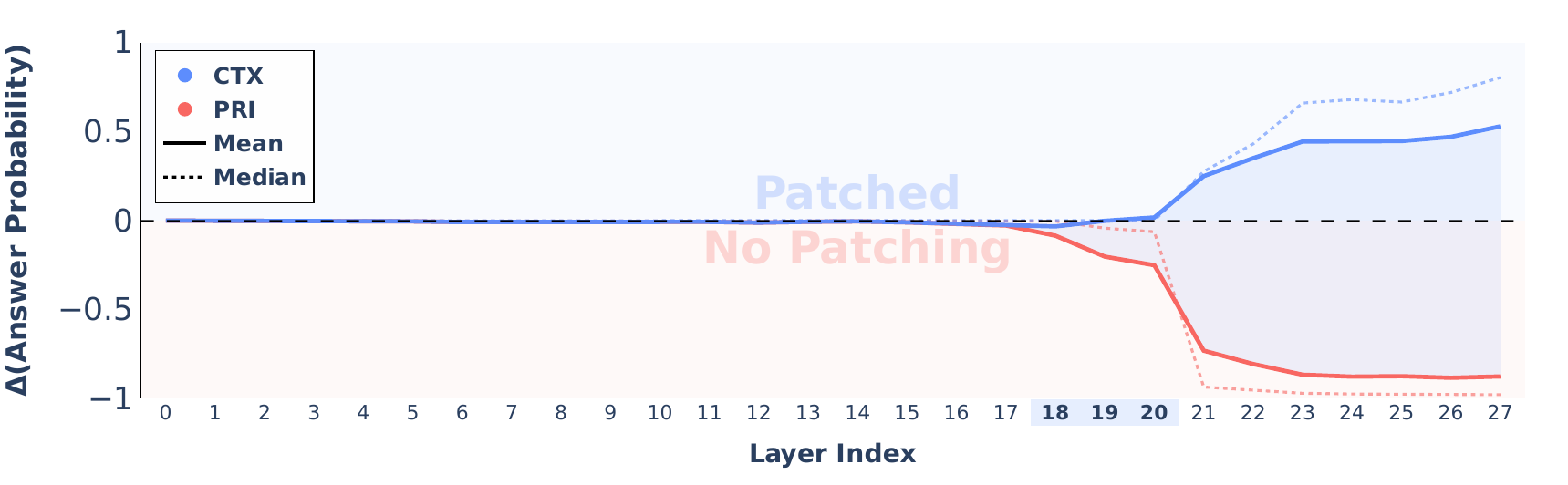}
\par\smallskip
Qwen2.5-VL-7B, ctx$\to$pri

\includegraphics[width=0.8\linewidth]{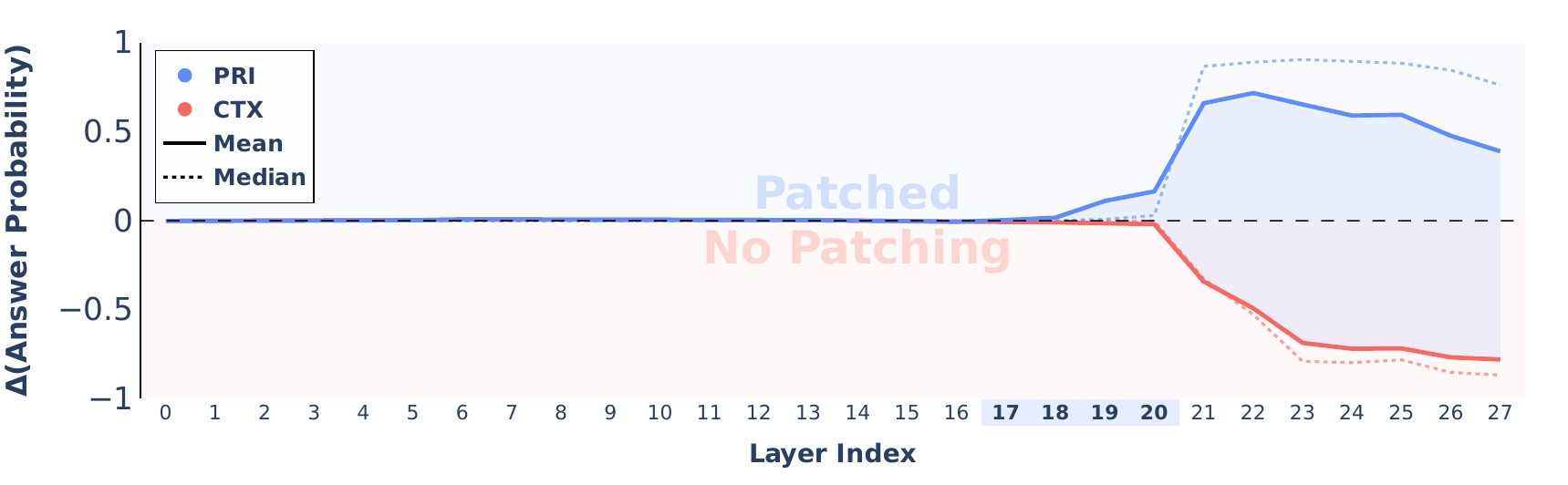}
\par\smallskip
Qwen2.5-VL-7B, pri$\to$ctx

\caption{\textbf{Cross-direction patching curves for the Qwen2.5-VL models (image channel).} Each pair shows both directions ($\ctxintent\!\to\!\priorintent$ and $\priorintent\!\to\!\ctxintent$). $\Delta$(answer probability) $=$ source $-$ target probability across layers; the shaded band marks the patched layers. These curves complement the main results in Figure~\ref{fig:patching_main}.}
\label{fig:patching_qwen25vl}
\end{figure}

\section{Comparison with Textual Context: Full Results}
\label{app:modality}
\begin{figure}[t]
    \centering
    \begin{subfigure}[b]{0.48\linewidth}
        \includegraphics[width=\linewidth]{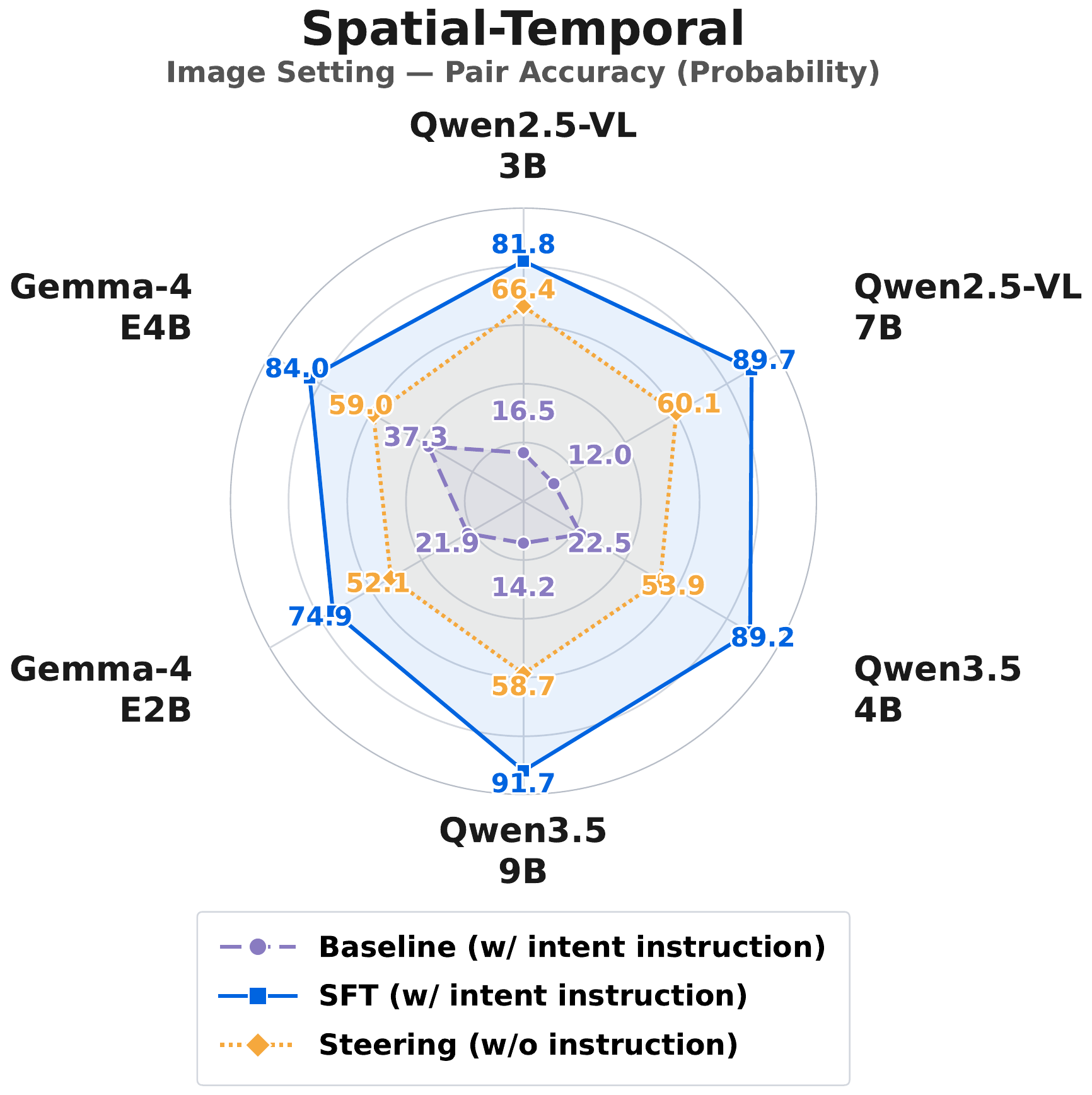}
    \end{subfigure}
    \hfill
    \begin{subfigure}[b]{0.48\linewidth}
        \includegraphics[width=\linewidth]{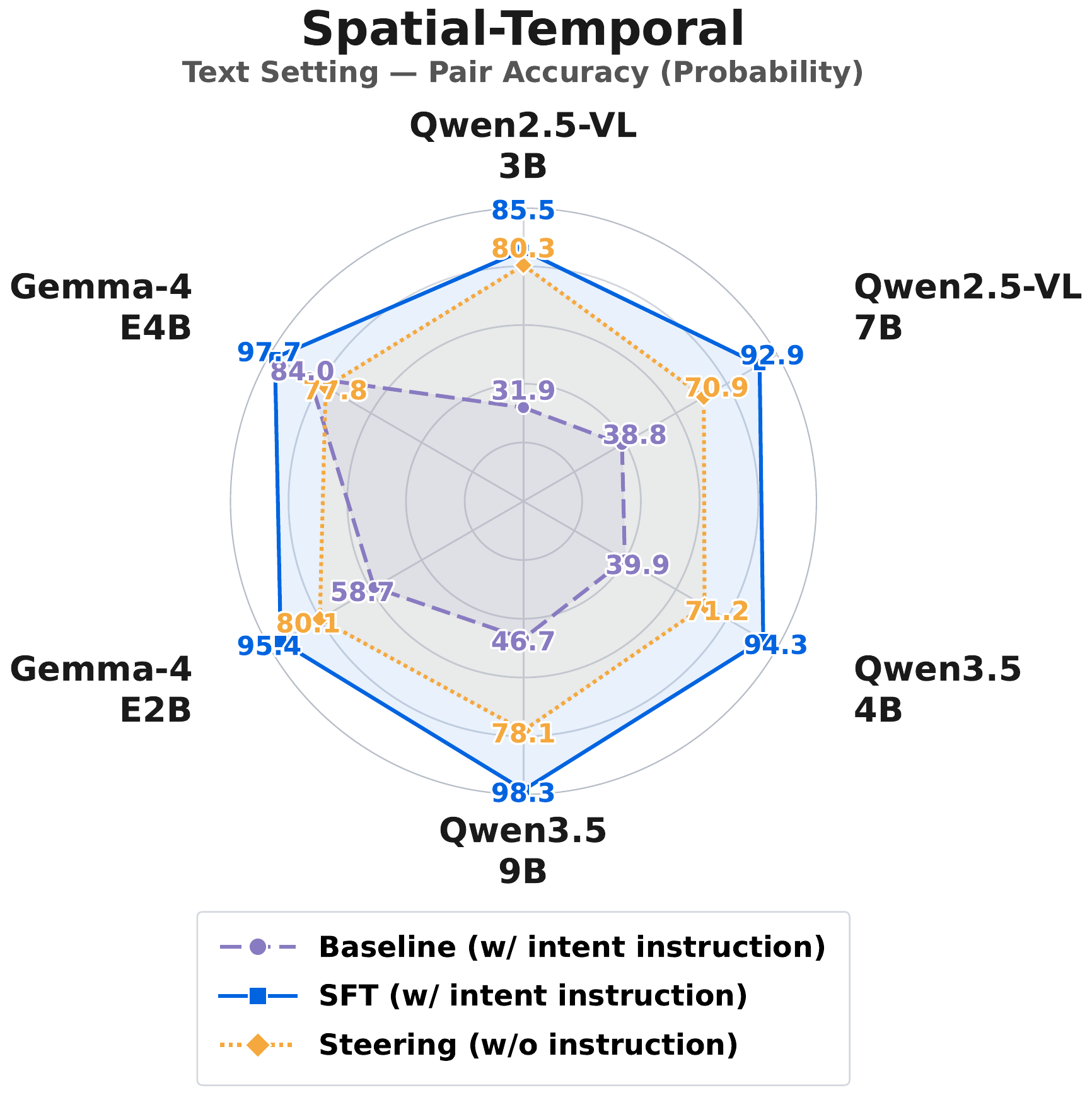}
    \end{subfigure}
    \caption{\textbf{Pair-accuracy on both channels for the Spatial-Temporal dataset.} Baseline, SFT, and steering results are shown for the image (left) and text (right) channels.}
    \label{fig:modality_spatial_temporal}
\end{figure}

\begin{figure}[t]
    \centering
    \begin{subfigure}[b]{0.48\linewidth}
        \includegraphics[width=\linewidth]{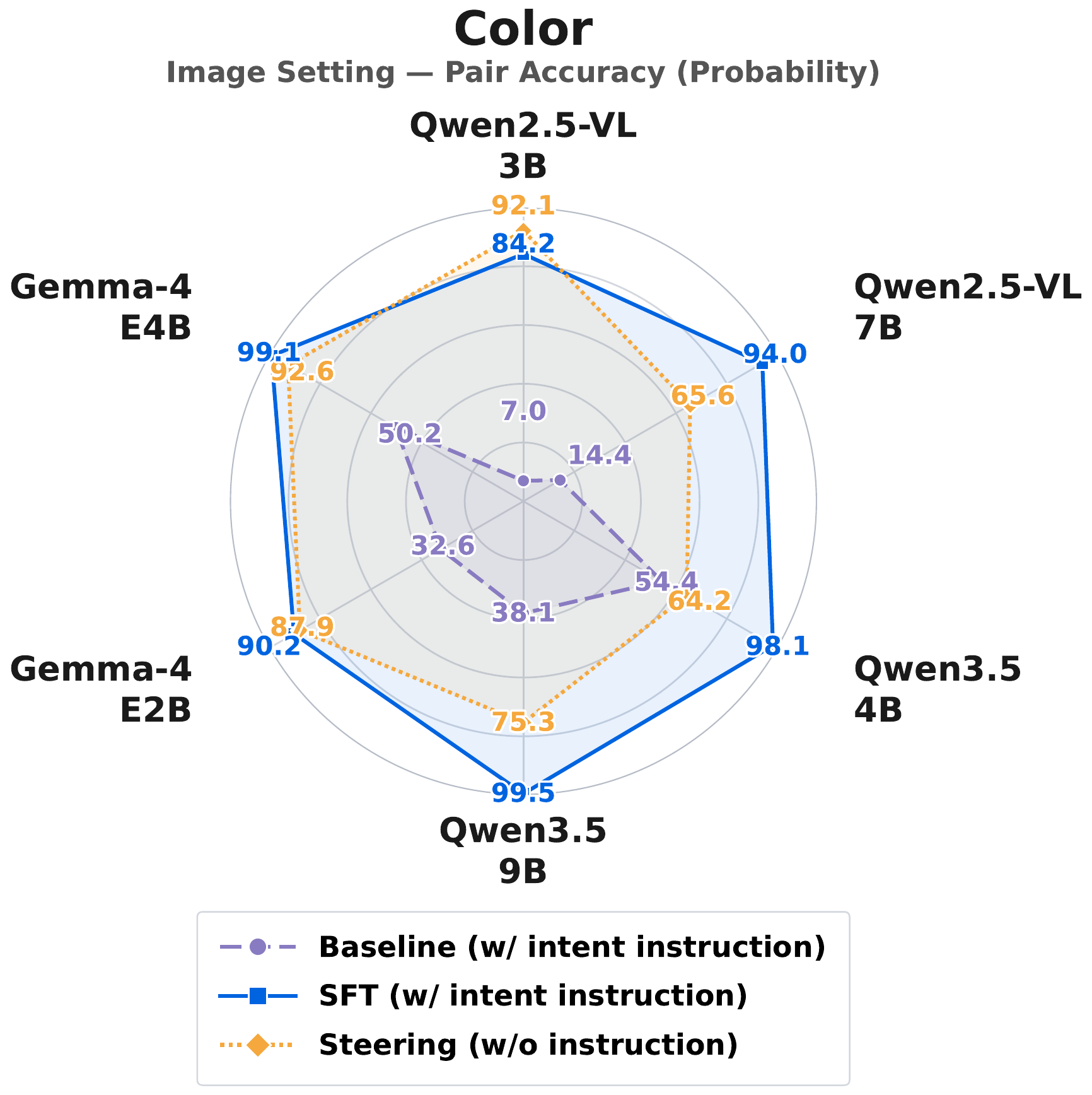}
    \end{subfigure}
    \hfill
    \begin{subfigure}[b]{0.48\linewidth}
        \includegraphics[width=\linewidth]{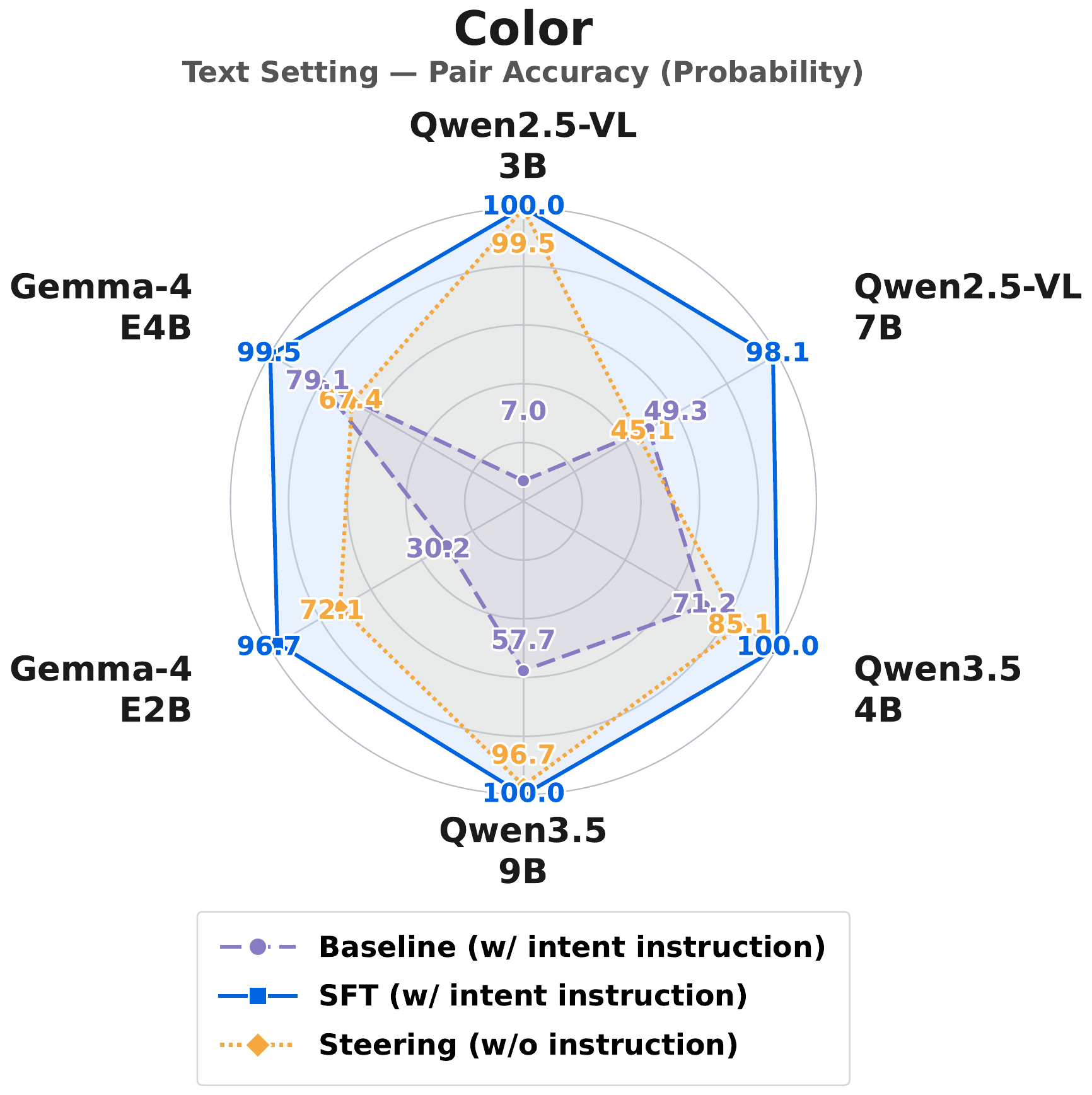}
    \end{subfigure}
    \caption{\textbf{Per-task pair-accuracy on both channels for the Color dataset.} Baseline, SFT, and steering per task are displayed.}
    \label{fig:modality_color}
\end{figure}

\begin{figure}[t]
    \centering
    \begin{subfigure}[b]{0.48\linewidth}
        \includegraphics[width=\linewidth]{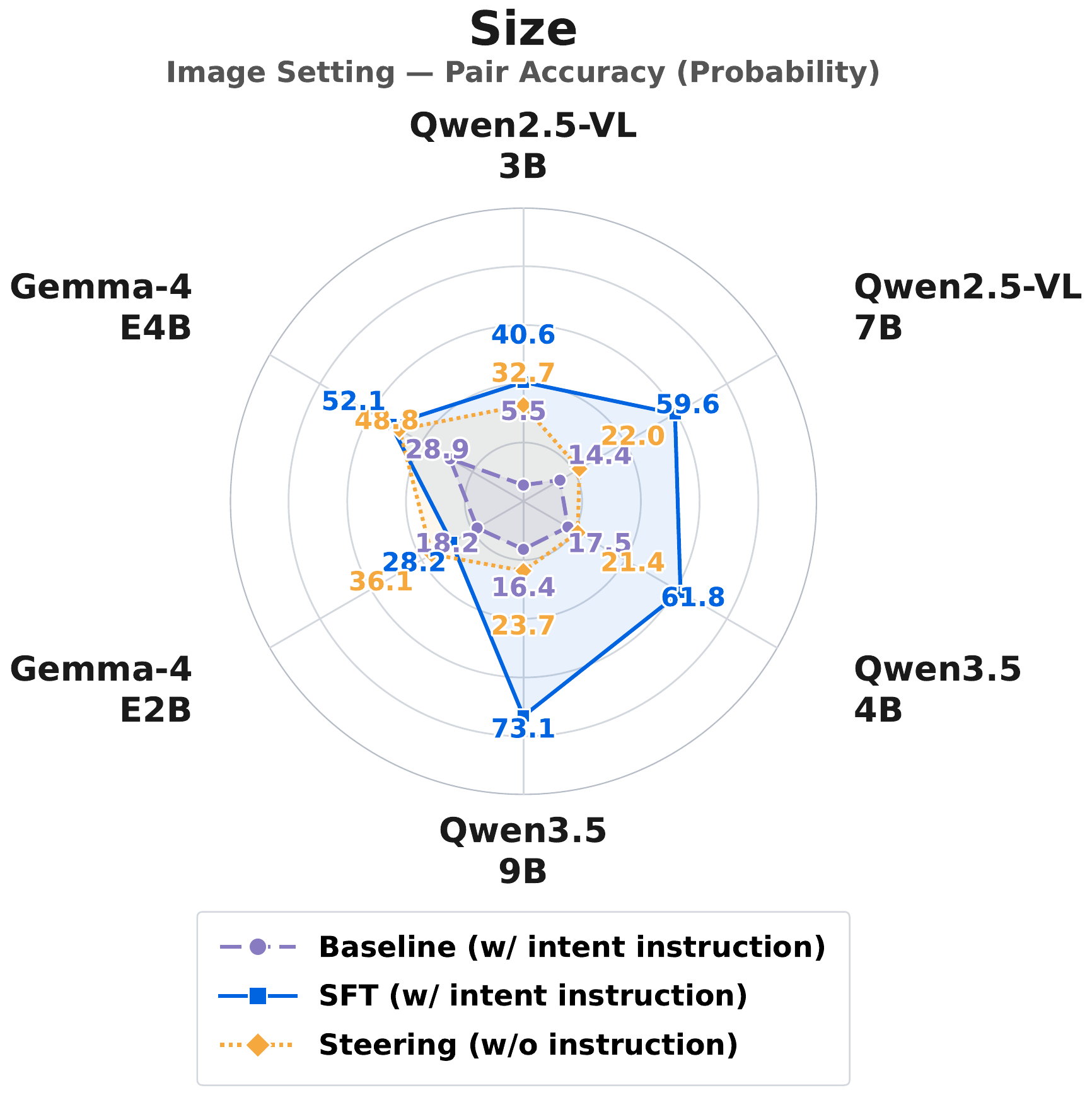}
    \end{subfigure}
    \hfill
    \begin{subfigure}[b]{0.48\linewidth}
        \includegraphics[width=\linewidth]{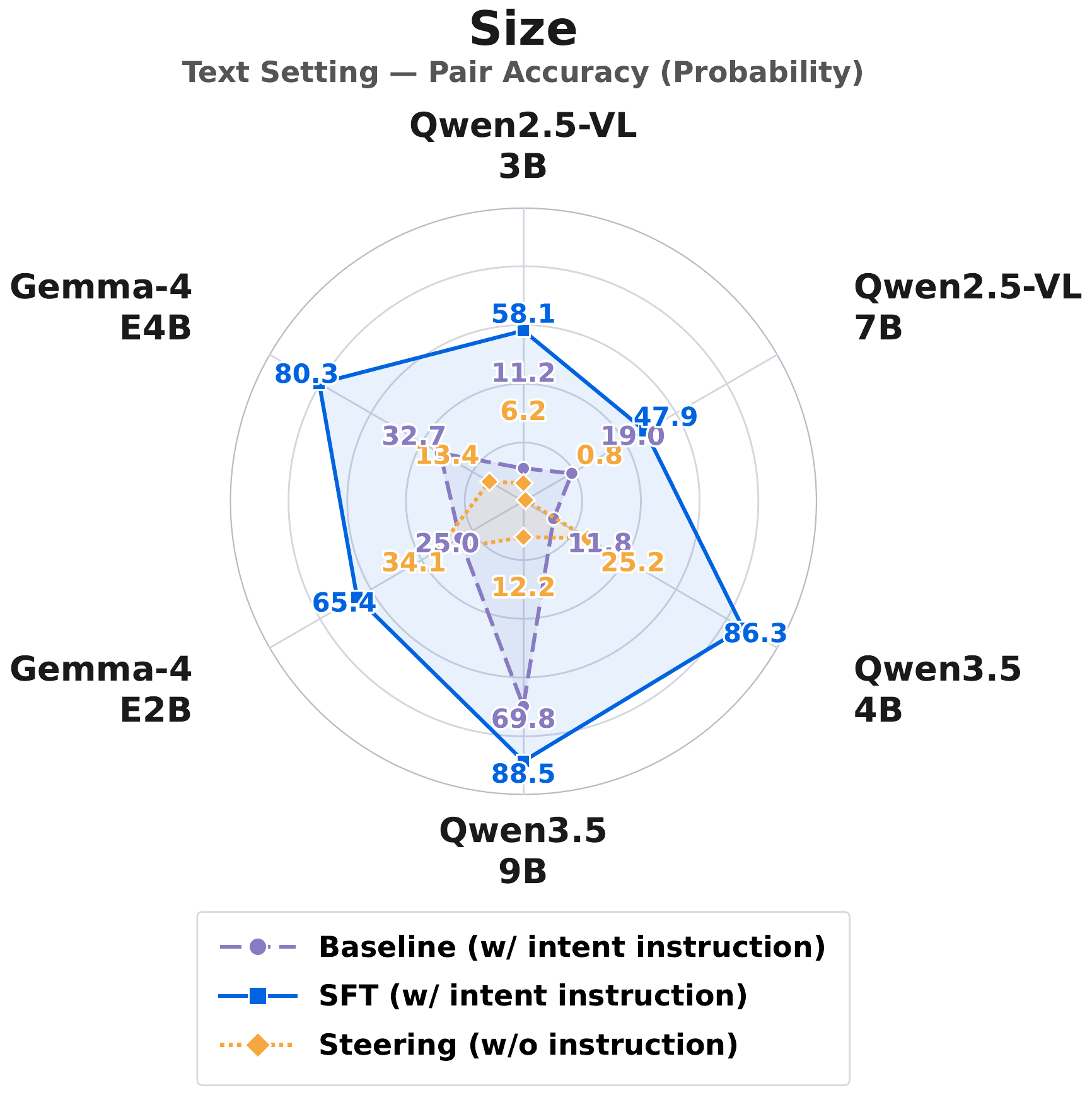}
    \end{subfigure}
    \caption{\textbf{Pair-accuracy on both channels for the Size dataset.} The image (left) and text (right) channels are compared across Baseline, SFT, and steering.}
    \label{fig:modality_size}
\end{figure}

\begin{figure}[t]
    \centering
    \begin{subfigure}[b]{0.48\linewidth}
        \includegraphics[width=\linewidth]{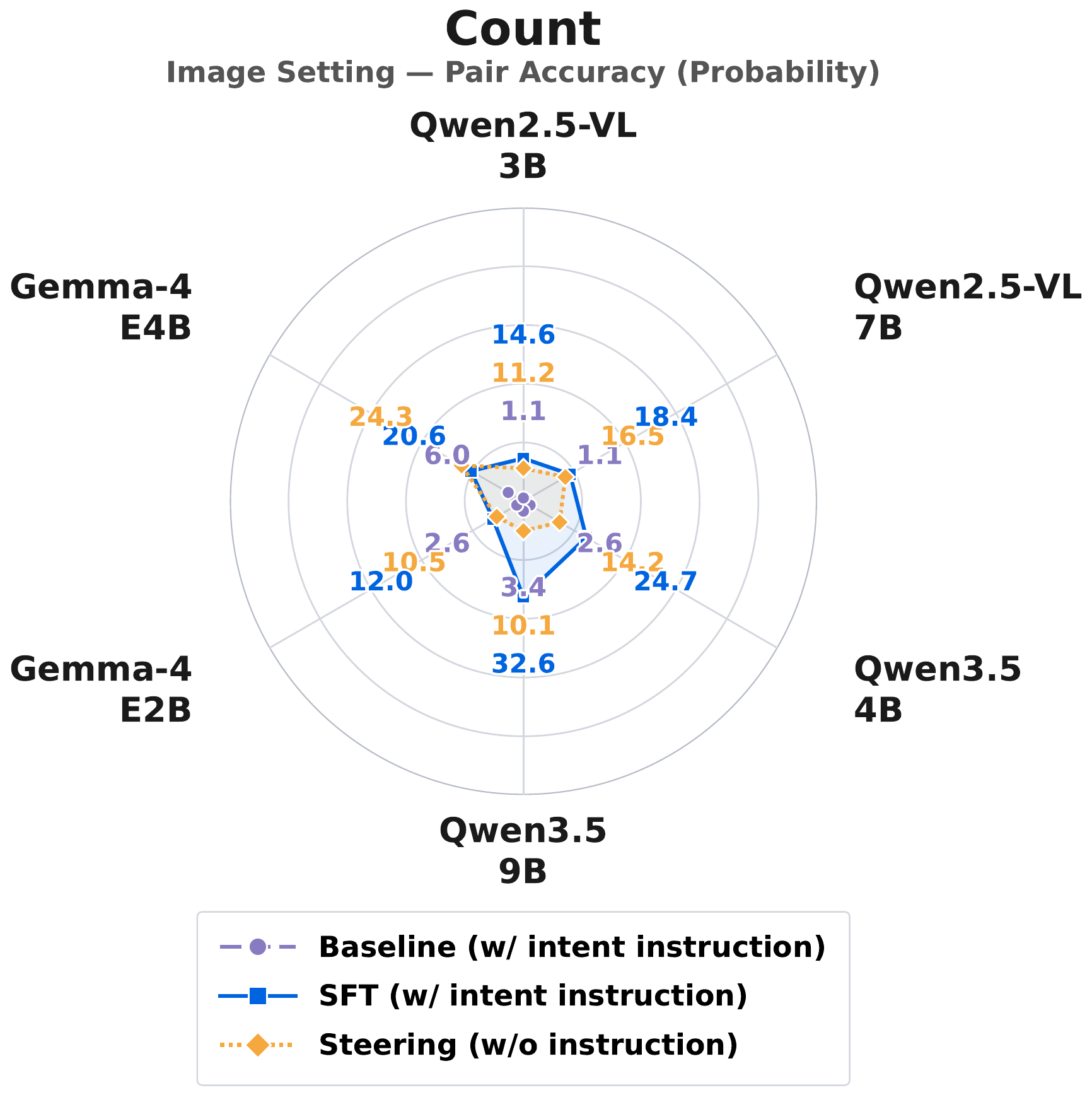}
    \end{subfigure}
    \hfill
    \begin{subfigure}[b]{0.48\linewidth}
        \includegraphics[width=\linewidth]{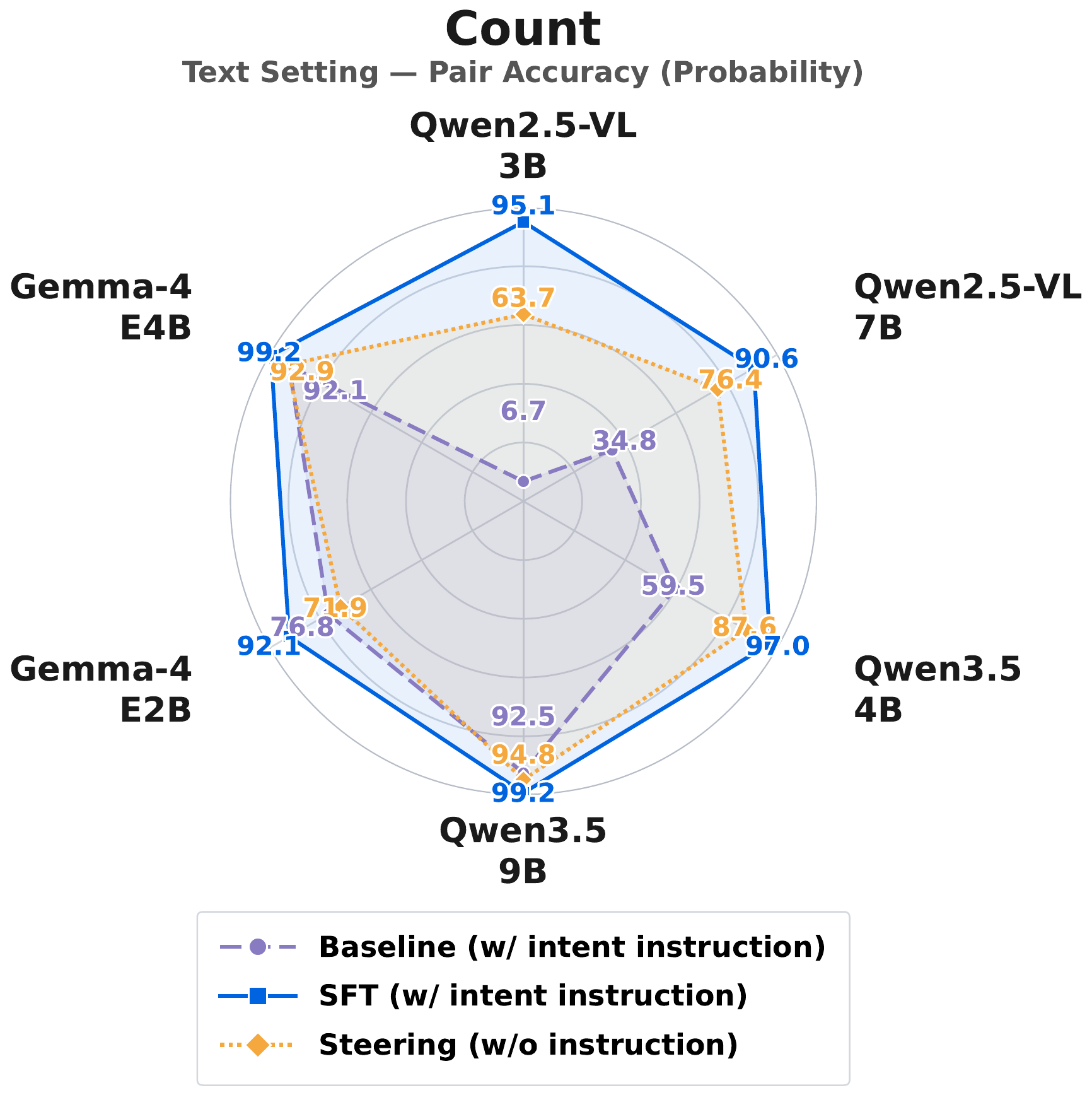}
    \end{subfigure}
    \caption{\textbf{Pair-accuracy on both channels for the Count dataset.} Baseline, SFT, and steering results are shown.}
    \label{fig:modality_count}
\end{figure}

\begin{figure}[t]
    \centering
    \begin{subfigure}[b]{0.48\linewidth}
        \includegraphics[width=\linewidth]{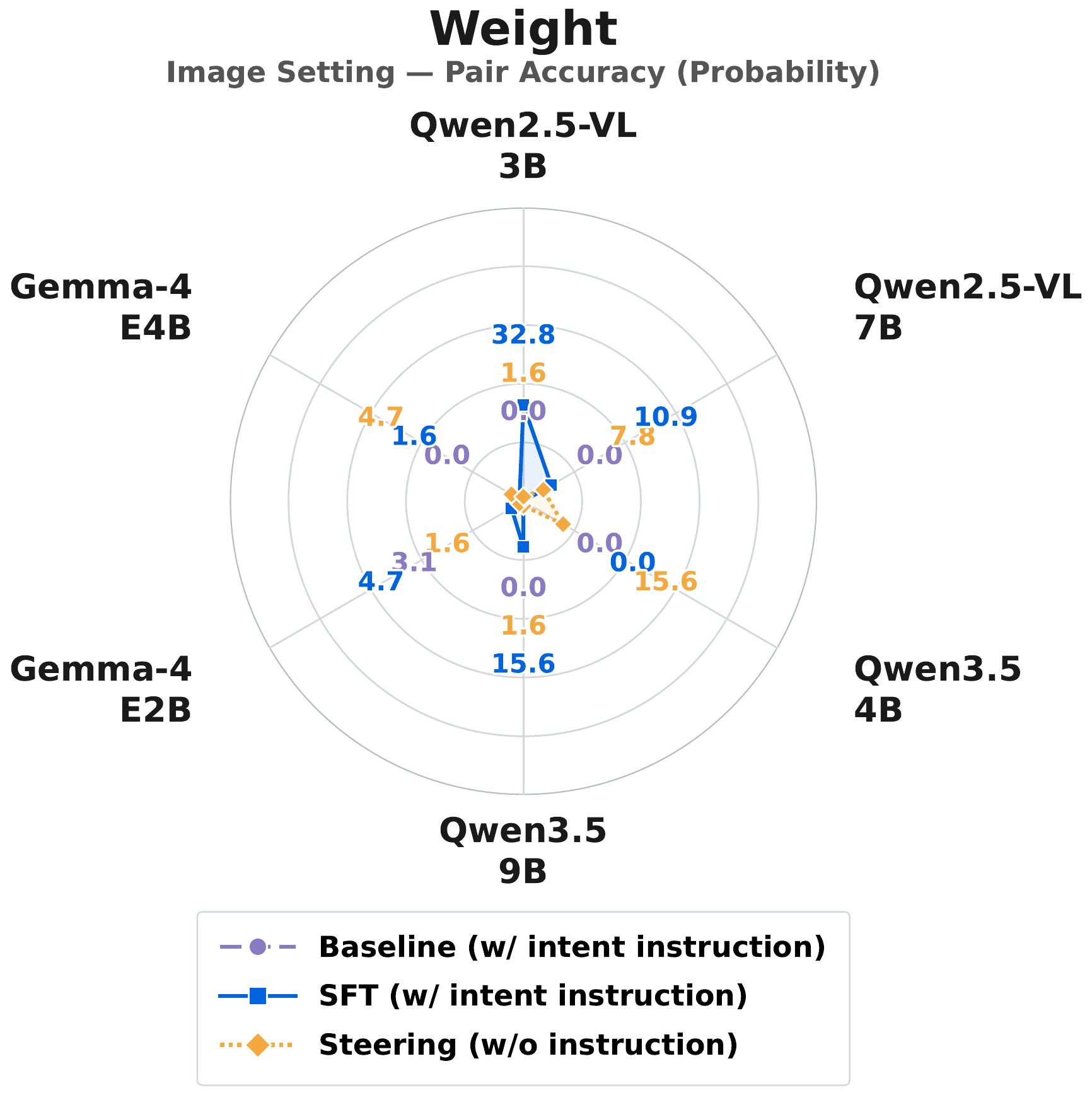}

    \end{subfigure}
    \hfill
    \begin{subfigure}[b]{0.48\linewidth}
        \includegraphics[width=\linewidth]{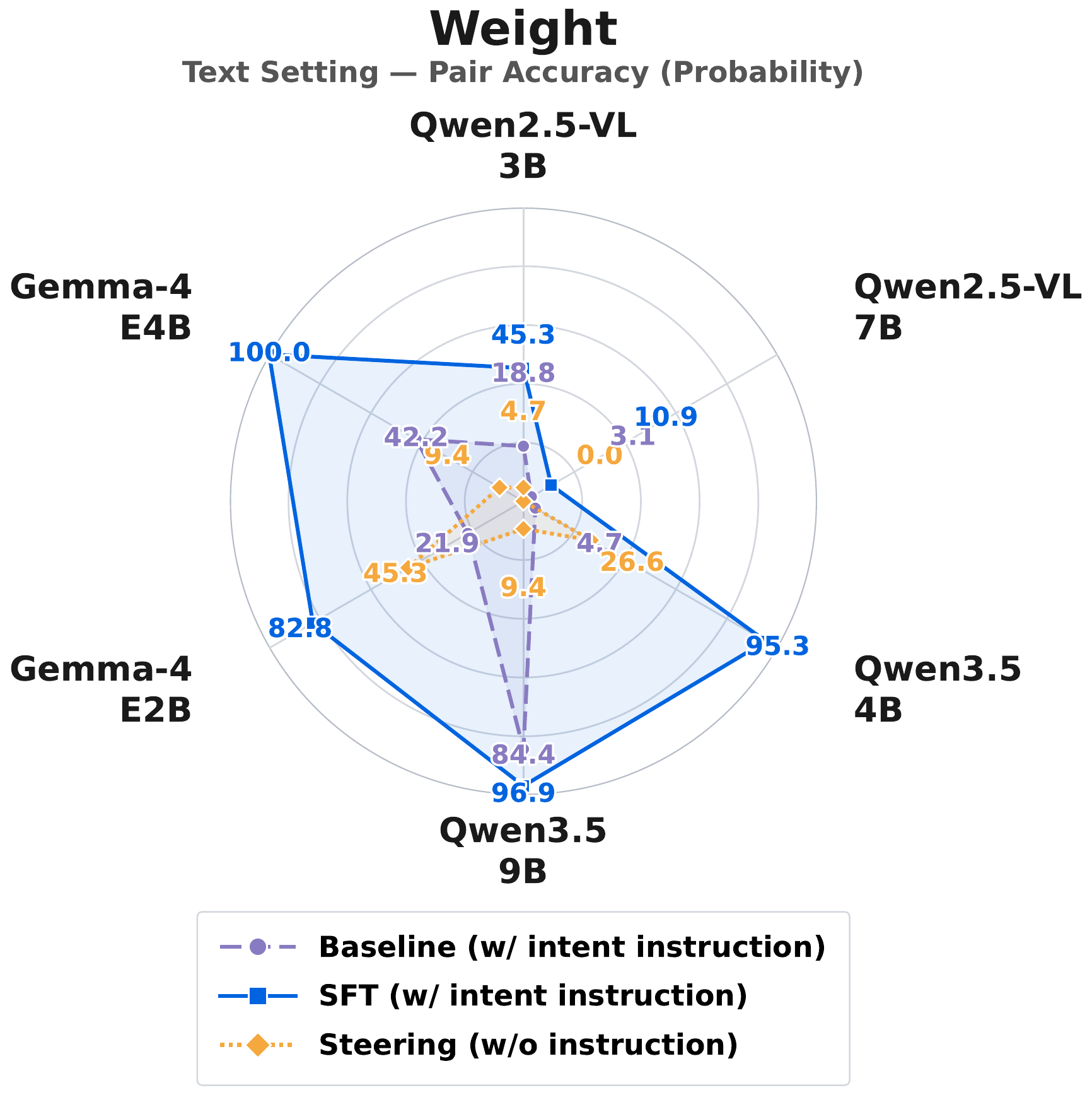}

    \end{subfigure}
    \caption{\textbf{Pair-accuracy on both channels for the Weight dataset.} The image and text channels are compared for Baseline, SFT, and steering.}
    \label{fig:modality_weight}
\end{figure}

Section~\ref{sec:modal_comparison} presents the modality comparison; here we give the supporting tables, per-task figures, and the cross-modality locus analysis. Every counterfactual image in \benchmark is paired with a one-sentence statement asserting the same fact, so the text channel runs the identical pipeline (base, SFT, steering) on matched content. Because the statement asserts the counterfactual outright, the text condition is an instruction-following \emph{upper bound}, not a fair head-to-head; we use it only as a same-mechanism reference for how much harder the visual channel is to control. Table~\ref{tab:modality_gap} reports the macro pair-accuracy of both channels under each condition (base $14.7\%$ image vs.\ $43.4\%$ text; SFT widens the gap to $33.6$ pp; steering narrows it to $15.6$ pp), and Figure~\ref{fig:modality_spatial_temporal}, \ref{fig:modality_color}, \ref{fig:modality_size}, \ref{fig:modality_count} and \ref{fig:modality_weight} breaks the two channels down per task.

\rparagraph{The per-task gap tracks visual difficulty.} On Color and Spatial-Temporal, vanilla models already achieve 32.8\% and 20.8\% in the image-only setting, approaching the text-only scores of 49.1\% and 50.0\%. In contrast, on Count and Weight, image-only performance sits near the floor (2.8\% and 0.5\%), whereas text-only models perform strongly (60.4\% and 29.2\%). This divergence arises because counterfactual facts can be trivially stated in language but must be exhaustively enumerated or inferred from pixels (Figure~\ref{fig:modality_spatial_temporal}, \ref{fig:modality_color}, \ref{fig:modality_size}, \ref{fig:modality_count} and \ref{fig:modality_weight}).
This is the same attribute ordering seen in the SFT and steering lifts, and it points to the visual \emph{access} to the evidence, not the downstream decision, as what varies across tasks.

\end{document}